\documentclass[sigconf]{acmart}
\usepackage{amsmath,amssymb,amsfonts}
\usepackage{algorithmic}
\usepackage{algorithm}
\usepackage{graphicx}
\usepackage{textcomp}
\usepackage{xcolor}
\usepackage{microtype}
\usepackage{tabularx}
\usepackage{graphicx}
\usepackage{booktabs}
\def\BibTeX{{\rm B\kern-.05em{\sc i\kern-.025em b}\kern-.08em
    T\kern-.1667em\lower.7ex\hbox{E}\kern-.125emX}}

\begin{document}
\fancyhead{}

\title{SparseRT: Accelerating Unstructured Sparsity on GPUs for Deep Learning Inference}

\author{Ziheng Wang}
\email{ziheng@mit.edu}
\affiliation{%
  \institution{Department of Computer Science, Massachusetts Institute of Technology}
  \streetaddress{32 Vassar St}
  \city{Cambridge}
  \state{MA}
  \postcode{02139}
}

\begin{abstract}
In recent years, there has been a flurry of research in deep neural network pruning and compression. Early approaches prune weights individually. However, it is difficult to take advantage of the resulting unstructured sparsity patterns on modern hardware like GPUs. As a result, pruning strategies which impose sparsity structures in the weights have become more popular. However, these structured pruning approaches typically lead to higher losses in accuracy than unstructured pruning. In this paper, we present SparseRT, a code generator that leverage unstructured sparsity to accelerate sparse linear algebra operations in deep learning inference on GPUs. For 1x1 convolutions and fully connected layers, we demonstrate geometric mean of speedups of 3.4x over the equivalent dense computation at 90\% sparsity and 5.4x at 95\% sparsity when evaluated on hundreds of test cases in deep learning. For sparse 3x3 convolutions, we show speedups of over 5x on use cases in ResNet-50. 
\end{abstract}



\maketitle

\section{Introduction}

Ever since Han's seminal work on model compression and pruning \cite{han2015deep}, there has been significant research interest on how to compress deep neural networks (DNN). The objective is to significantly reduce the number of parameters in a DNN to reduce the inference overhead and memory footprint, while not losing too much accuracy. Many have demonstrated that if weights are pruned individually by magnitude or by some other criterion (variational dropout \cite{molchanov2017variational}, L0 norm \cite{louizos2017learning}), popular DNNs in computer vision and natural language processing can be pruned by upwards of 90\% with almost no loss in accuracy \cite{gale2019state,han2015deep}. 

Unfortunately, it was soon recognized that unstructured sparsity patterns cannot take advantage of modern general purpose hardware architectures such as multicore CPUs and GPUs typically used for deep learning inference. This discovery led to the developments of custom hardware \cite{han2016eie} as well as ``structured'' pruning methods that prune blocks of weights at once. The resulting weights from these structured pruning methods often can be used directly in dense BLAS routines or have enough structure to support highly performant implementations on current hardware \cite{gray2017gpu,yao2019balanced}. However, these structured pruning methods often lead to larger accuracy losses than unstructured pruning and do no better than a smaller dense network \cite{closerlook2019}.

The big accuracy losses resulting from structured pruning led us to reexamine to what extent unstructured sparsity can be efficiently implemented with modern general purpose hardware architectures. In a DNN pruned via unstructured pruning, the core operation at inference time is typically \textbf{sparse matrix -- dense matrix multiplication} (commonly abbreviated as \textbf{SpMM}). This routine is at the core of the popular 1x1 convolutions (aka contractions) and fully connected layers. In latest computer vision network architectures such as MobileNet and EfficientNet, 1x1 convolutions typically represent the bulk of the parameters and FLOPs \cite{elsen2019fast,howard2017mobilenets,tan2019efficientnet}. Fully connected layers account for nearly all of the parameters and much of the FLOPs in deep recurrent or self-attention based models in natural language processing (NLP)\cite{devlin2018bert,wang2019structured}. Moreover, other important operations, such as sparse 3x3 convolutions, can be cast as a SpMM problem through the same techniques used in the dense case, such as im2col \cite{chetlur2014cudnn}. 

In accordance with general observations from practitioners, we find existing SpMM implementations to be poorly suited for accelerating unstructured sparse DNNs \cite{deepbench}. Indeed, current SpMM solutions are often catered to scientific computing applications where the sparse matrix dimensions are massive compared to the dense matrix dimension, and only achieve significant speedup when the sparsity ratio is very high (typically on the order of 99\% to 99.9\%) \cite{yang2018design,hong2019adaptive,hong2018efficient}. Typically, pruned neural networks can at most have 95\% sparsity without significant accuracy losses. In addition, in SpMM problems in deep learning, the dense matrix dimensions can be larger than the sparse matrix dimensions. 

In this work, we present SparseRT, a code generator for SpMM and sparse convolution kernels that are well suited for the deep learning inference case based on the inspector-executor optimization framework with a PTX code generation backend. In SparseRT, instead of trying to cater to the idiosyncrasies of sparse matrix formats, we start from a dense matrix multiplication and simply remove unnecessary computations in an inspector-executor framework \cite{venkat2015loop,intelieblas}.  When evaluated on the latest GPUs, we show that our strategy can achieve a geometric mean of 3.4x speedup on SpMM with 90\% sparsity and 5.4x speedup at 95\% sparsity on hundreds of test matrices from pruned neural networks in computer vision and NLP. Lowering sparse convolution to SpMM, we are able to achieve up to 4x speedup on the 3x3 sparse convolutions in ResNet-50. We hope this work puts in question the conventional wisdom that unstructured sparsity is poorly supported on commodity parallel hardware architectures and inspires future research on unstructured pruning of neural networks.

In this work, we only focus on improving the performance of key primitives, SpMM and sparse convolution, that accelerate single layers in a sparse DNN. We do not explore inter-layer optimizations such as operator fusion and concurrent scheduling, used in state-of-the-art frameworks such as TensorRT \cite{tensorrt}. As a result, we only benchmark performance of single layers to avoid these confounding factors in this paper. Incorporating these optimizations into SparseRT is important future work. 

\section{Background}
This section briefly discusses computations that arise in neural network pruning and current attempts to handle those computations on commodity hardware.

\subsection{Neural Network Pruning}

The large size of neural networks is a major obstacle in their deployment in production. The hundreds of millions of FLOPs required to run an input example through the DNN lead to severe latency issues. Fortunately, Han et al. established that modern DNNs are heavily overparameterized. Up to 90\% of the weights can be eliminated with virtually no loss in accuracy \cite{han2015deep}. However, the pruning method proposed by Han et al. results in unstructured sparse weight matrices, which are not well-supported by commonly used hardware like GPUs. The poor fit between unstructured sparsity and modern general purpose hardware has given rise to many research efforts in structured pruning techniques. In convolution operations, entire channels are pruned at once from filters \cite{he2017channel}. In recurrent neural networks, entire neurons are removed from the network \cite{wen2017learning}. These structured pruning approaches directly speed up inference as they replace the original dense weight matrix with a smaller dense weight matrix. Highly optimized dense linear algebra primitives can directly be used to obtain good inference performance. 

However, recent work has shown that most structured pruning techniques often result in unacceptably large accuracy loss: one can often achieve better performance by simply training a smaller dense neural network from scratch \cite{closerlook2019}. This result motivated us to re-examine the assumption that unstructured sparsity cannot be efficiently implemented on modern hardware architectures such as GPUs. 

\subsection{SpMM Performance}

Parallel to the recent model pruning research efforts in the deep leaning community, the high performance computing community has been working to support sparse linear algebra computations on modern parallel hardware architectures. Good performance has been achieved for SpMM operations frequently encountered in scientific computing applications, where the dense matrix dimensions are highly skewed: it resembles a small collection of vectors (multi-vector) rather than a matrix \cite{hong2019adaptive,hong2018efficient}. However, the SpMM operations in deep learning might have quite balanced dimensions. The sparse matrix may even be smaller than the dense matrix. Existing SpMM implementations often fail to bring speedups in these cases. For example, SpMM operations in Deepbench where the dimensions are relatively balanced can be 10 times slower than the dense operation on recent GPUs like Titan Xp with vendor library cuSPARSE\cite{deepbench}. This is because current SpMM implementations are often just extensions of sparse matrix vector multiplication (\textbf{SpMV}) with limited tiling over the dense matrix dimension, resulting in unfavorable linear cost scaling in the dense matrix dimension.

Recent works have sought to address this challenge. Elsen et al. proposed efficient kernels for SpMM on mobile platforms, and showed that unstructured sparsity can be exploited to improve performance by as much as 2x \cite{elsen2019fast}. In parallel, there has been recent works exploring sparse convolutions \cite{baghdaditiramisu,chen2019escoin,park2016faster}. While dense convolutions can make use of intricate optimizations such as the Winograd transform \cite{lavin2016fast}, sparse convolutions typically rely on direct convolution \cite{chen2019escoin}, or implicit GeMM based approaches that cast the problem as a SpMM \cite{baghdaditiramisu,park2016faster}. In this work, we adopt the latter approach to support sparse convolutions.

\subsection{Inspector-Executor Framework}

A common strategy to optimize sparse matrix computations is the inspector-executor framework \cite{venkat2015loop,strout2018sparse,intelieblas,strout2004sparse}. In this framework, to perform a sparse operation, the sparse matrix input is analyzed in an inspector phase to generate optimized code tailored to its sparsity pattern and dimensions. The optimized code (``executor'') is then executed to perform the computation. The inspector phase is typically done at runtime, and its efficiency is of significant concern \cite{venkat2015loop}. However, in deep learning inference, it can be done offline at compile time as the sparse weight matrix is fully known. It does not impact the runtime inference latency.

For each SpMM or sparse convolution problem tested, SparseRT inspects the sparse weight matrix statically and enumerate optimized executors corresponding to different code transformation strategies. We then autotune for the executor with the best performance. Similarly, the autotuning is done at compile time, and has no effect on the runtime inference latency.

\subsection{Autotuning}

The concept of autotuning has been around for a long time \cite{ansel2014opentuner,whaley2001automated}. High performance computing kernels almost always have several parameters, like tile size and loop order, that can be adjusted. Different settings of these parameters lead to different performance characteristics and runtimes of the kernel. In applications where there are no strict requirements on the compilation time, such as deep learning inference, one can search through these different settings to find the one with the best performance (aka tuning the kernel). Recently, a few different end-to-end systems have been proposed to perform autotuning on individual kernels and entire neural networks to achieve good inference performance \cite{tillet2017input,tillet2019triton,chen2018tvm,vasilache2018tensor}. However, all these recent systems currently only focus on dense neural networks.

\section{Design Principles for Fast SpMM}

In this section, we first briefly describe the CUDA programming model and relevant terminology. As aforementioned, SparseRT follows the inspector-executor framework. At compile time, the sparse weight matrix is fully known. The sparse matrix is first treated as a dense matrix, casting the problem to a GeMM. We then explore different tiling strategies for the GeMM problem. For each of those tiling strategies, we rebalance and remove computations that involve zeros in the original sparse matrix to obtain an optimized executor code in PTX or CUDA. Finally, we autotune for the fastest executor.

\subsection{CUDA Programming Model}

In this work, we use the CUDA programming model for all implementations. While a detailed treatise can be found elsewhere, we briefly introduce key terminology and principles. In CUDA, work in a program is divided among processes named threads. Each thread has private \textit{registers} which can be quickly accessed, as long as not too many registers are used overall. Efficient usage of those registers is often critical to achieving good performance. Besides registers, GPUs offer a \textit{constant cache}, which allows extremely efficient access of compile time constants. Importantly, the constant cache is shared with instruction cache in latest GPU architectures \cite{jia2019dissecting} and does not cause cache contention with on-chip data caches. 

Threads can be divided into \textit{thread groups} (aka cooperative groups \cite{harris2017cooperative}). A thread group is a useful construct for GPU performance as GPU hardware strongly prefer that a \textit{warp} of 32 threads execute the same instruction per cycle. A thread group can contain one of more warps. Different thread groups can thus execute different instruction streams with little to no performance overhead, in contrast to the severe warp divergence penalties that would ensue if threads within a warp execute different instruction streams. 

Multiple thread groups can be grouped together to form a \textit{thread block}, which also offers a scratchpad memory space called \textit{shared memory} with which the thread groups can communicate. A kernel launched on the GPU typically consists of several such thread blocks. There are very limited synchronization options among different thread blocks. 

Typically, high performing GPU kernels are \textit{load-balanced}. This means that different thread groups have roughly the same amount of work, since a thread block cannot be retired from the device until all thread groups finish. Similarly, all thread blocks should roughly have the same amount of work, since a kernel launch cannot terminate until all thread blocks finish execution. In sparse operations, typically the amount of work owned by a thread block is proportional to the number of nonzeros it has to process. As a result, different thread blocks should be assigned similar number of nonzeros in the sparse matrix.

\subsection{Terminology}

Here we introduce the terminology we will use to describe matrix multiplication. Our discussion will be based on the matrix multiplication $A \times B = C$, where matrix $A$ has dimensions $M \times K$, $B$ has dimensions $K \times N$ and $C$ has dimensions $M\times N$. To compute the matrix multiplication, we iterate over 3 axes, M, K and N. Note that when M, K and N are not italicized, they refer to an iteration domain. When they are, they refer to specific dimensions (the size of the iteration domain). So for example, axis M has $M$ elements. We call K the \textit{reduction} axis and M, N the \textit{external} axes. In SpMM, we fix the sparse matrix to be $A$ and the dense matrix to be $B$. Unlike cuBLAS, in the following discussion, we assume a C-style row-major default layout of the matrices. We assume that $A$ is transposed. (In the DNN inference setting, the data layout of the weights can be picked arbitrarily to maximize performance.)

\subsection{Densify and Tile}

\begin{figure}[t]
\begin{center}
\includegraphics[width=1\linewidth]{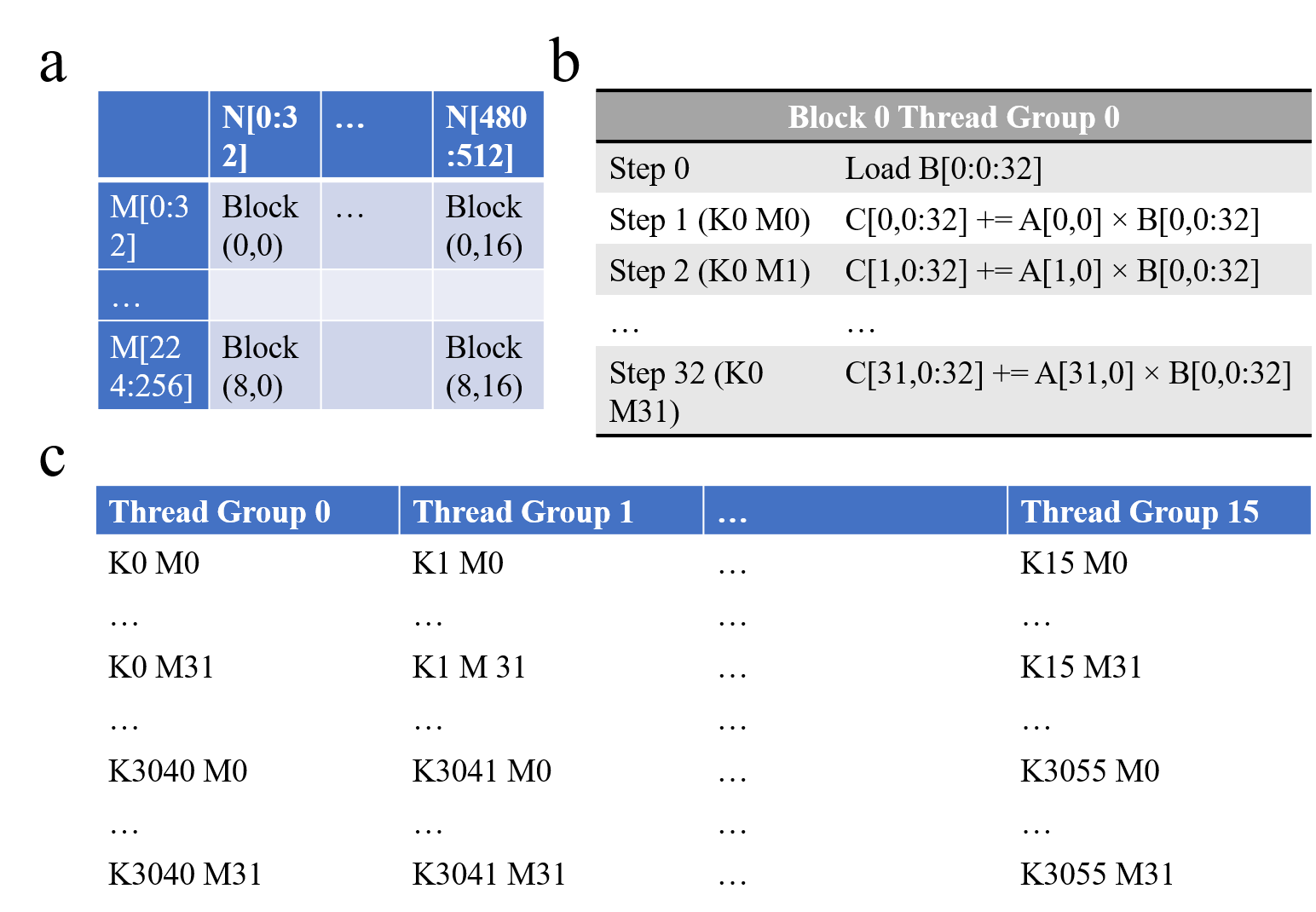}
\end{center}
   \caption{Tiling Strategy for GeMM ($M=512,K=3056,N=256$). a) Each block in the 2-D grid of blocks processes a rectangular tile of the output. b) Thread group computation for one B element, corresponding to lines 8-11 in Algorithm 2. c) Tiling strategy at granularity of K and M. Each thread group processes a range of M indices for each K index.}
\label{fig:tiling}
\end{figure}

The first step in SparseRT's handling of a SpMM problem is treating it as a dense GeMM problem. We then tile this GeMM problem. Here, we explain our parameterizable tiling strategy in the context of an example problem, $M=256,K=3056,N=512$. We split the external axes over a two dimensional grid of thread blocks of size ($M\_blocks, N\_blocks$). Each thread block processes the entire reduction axis.  Each thread block is responsible for $N/N\_blocks$ N elements and $M/M\_blocks$ M elements. In this example illustration, $M\_blocks = 8, N\_blocks = 16$, so each thread block is responsible for processing 32 elements of the M and N axes each. This means that the thread block is responsible for producing a 32 $\times$ 32 portion of the output. For example, thread block (0,0) would produce $C[0:32,0:32]$. This is illustrated in Figure \ref{fig:tiling}a.

Each thread block is divided into $Gy$ thread groups of size $Gsy$ which process different portions of the reduction axis. In this example illustration, $Gy = 16$, so each thread group processes $3056/16 = 191$ elements of the reduction axis. Thread group 0 of thread block (0,0) would process $C[i,k] = \sum_{j=0}^{190}A[i,16j]\times B[16j,k]$, for $i\in\{0,1,...31\}$ and $j\in\{0,1,...31\}$. Each thread group holds an accumulation buffer of the 32 $\times$ 32 output tile in a register array distributed across threads. All accumulation buffers are reduced at the end of the thread block execution using shared memory. 

To execute this accumulation for one K element, the thread group maps one of the external indices, N, to the $Gsy$ threads in the thread group. The N elements are now tiled at tile size $Gsy$. We loop over the tiles in an inner loop of length $N/N\_blocks/Gsy$, with an outer loop of size $M/M\_blocks$ for all the M elements. In our work, we fix the inner loop size to 1, so $Gsy$ is fixed to be $N/N\_blocks$. This is depicted in Figure \ref{fig:tiling}b. The thread group first loads a vector of $B$ elements, and then loops through a vector of $A$ elements. Figure \ref{fig:tiling}c illustrates the tiling strategy at the granularity of iterations of K and M indices. The pseudocode for the example is listed in Algorithm \ref{alg:example}, while the general case is listed in Algorithm \ref{alg:generic}.

\begin{algorithm}[tb]
   \caption{GeMM at thread group level: example}
   \label{alg:example}
\begin{algorithmic}[1]
   \STATE {\bfseries Input:} matrix pointers $A,B,C$
  
   \STATE $ACC[0:32,0:32] = \{0.0\} $ 
   \FOR{$b$ in $0,16,32...3040$}
   \STATE Cache $A[0:32,b]$ in shared memory
   \STATE Cache $B[b,0:32]$ in registers
    \FOR{$a$ in $0,1 ... 32$}
    \STATE $ACC[a,0:32] += A[a,b] \times B[b,0:32]$
    \ENDFOR 
    \STATE Add $ACC$ to $C$ atomically using shared memory
    \ENDFOR
\end{algorithmic}
\end{algorithm}

\begin{algorithm}[tb]
   \caption{GeMM at thread group level: general case}
   \label{alg:generic}
\begin{algorithmic}[1]
   \STATE {\bfseries Input:} matrix pointers $A,B,C$
   \STATE $K_{list}$ = list of K elements to process for thread group
   \STATE $M_{list}$ = list of M elements to process for thread group
   \STATE $N_{list}$ = list of N elements to process for thread group
   \STATE $ACC[M_{list},N_{list}] = \{0.0\} $
   \FOR{$b$ in $K_{list}$}
   \STATE Cache $A[M_{list},b]$ in shared memory
   \STATE Cache $B[b,N_{list}]$ in registers
    \FOR{$a$ in $M_{list}$}
    \STATE $ACC[a,N_{list}] += A[a,b] \times B[b,N_{list}]$
    \ENDFOR 
    \STATE Add $ACC$ to $C$ atomically using shared memory
    \ENDFOR
\end{algorithmic}
\end{algorithm}

We note that $M\_blocks, N\_blocks$ and $Gy$ are all tunable parameters of this algorithm.  

\begin{figure}[t]
\begin{center}
\includegraphics[width=1\linewidth]{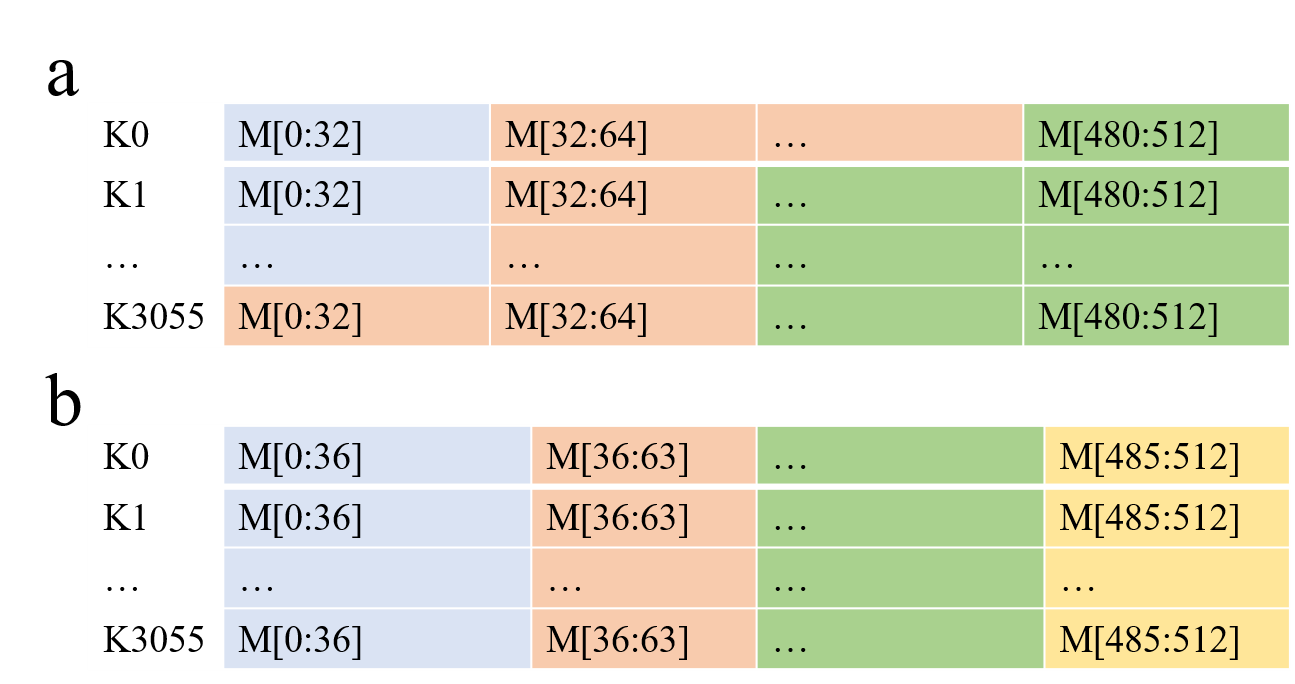}
\end{center}
   \caption{Thread block level load balancing strategies for SpMM ($M=512,K=3056,N=256$). Different colors denote assignment to different thread blocks a) Different thread blocks can have different portions of the reduction axis. b) Different thread blocks can have different portions of the external axis.}
\label{fig:balance}
\end{figure}

\subsection{Load Balancing}

To extend the above GeMM algorithm for SpMM, we first consider load balancing among thread blocks. In the dense GeMM algorithm, each thread block processes same-sized portions of the $A$ matrix (all elements of K and some elements of M). When $A$ is sparse, these same-sized portions might not contain the same amount of nonzeros. We see in principle two ways to perform the load balancing. The first way is to have different thread blocks process different number of K elements, while keeping the number of M elements processed by each thread block the same. (Figure \ref{fig:balance}a) The second way is to assign different number of elements in M to different thread blocks to balance the number of nonzero values in the thread blocks. (Figure \ref{fig:balance}b)

Each strategy has its own benefits and drawbacks. The first method allows for finer grained load balancing. However, each thread block might process more than one output tile if a thread block is assigned over two columns, as indicated in Figure \ref{fig:balance}a. This leads to more complicated logic in the generated code and atomic writes to global memory. The second strategy ensures that each thread block only processes one output tile, though the tiles might be of different sizes. This means that the accumulator register array will be of different size in different thread blocks. The current CUDA programming model does not support allocation of different number of registers to different thread blocks. This means that some registers will be wasted to support the maximum sized register array required. The load balancing is also coarser than the first method, especially if $B$ is much larger than $A$. In practice, we adopt the second strategy (Figure \ref{fig:balance}b). We find that for the tested sparse matrices in deep learning, the sparsity pattern does not lead to badly skewed register usage among thread blocks or poor load balancing.

\begin{figure}[t]
\begin{center}
\includegraphics[width=1\linewidth]{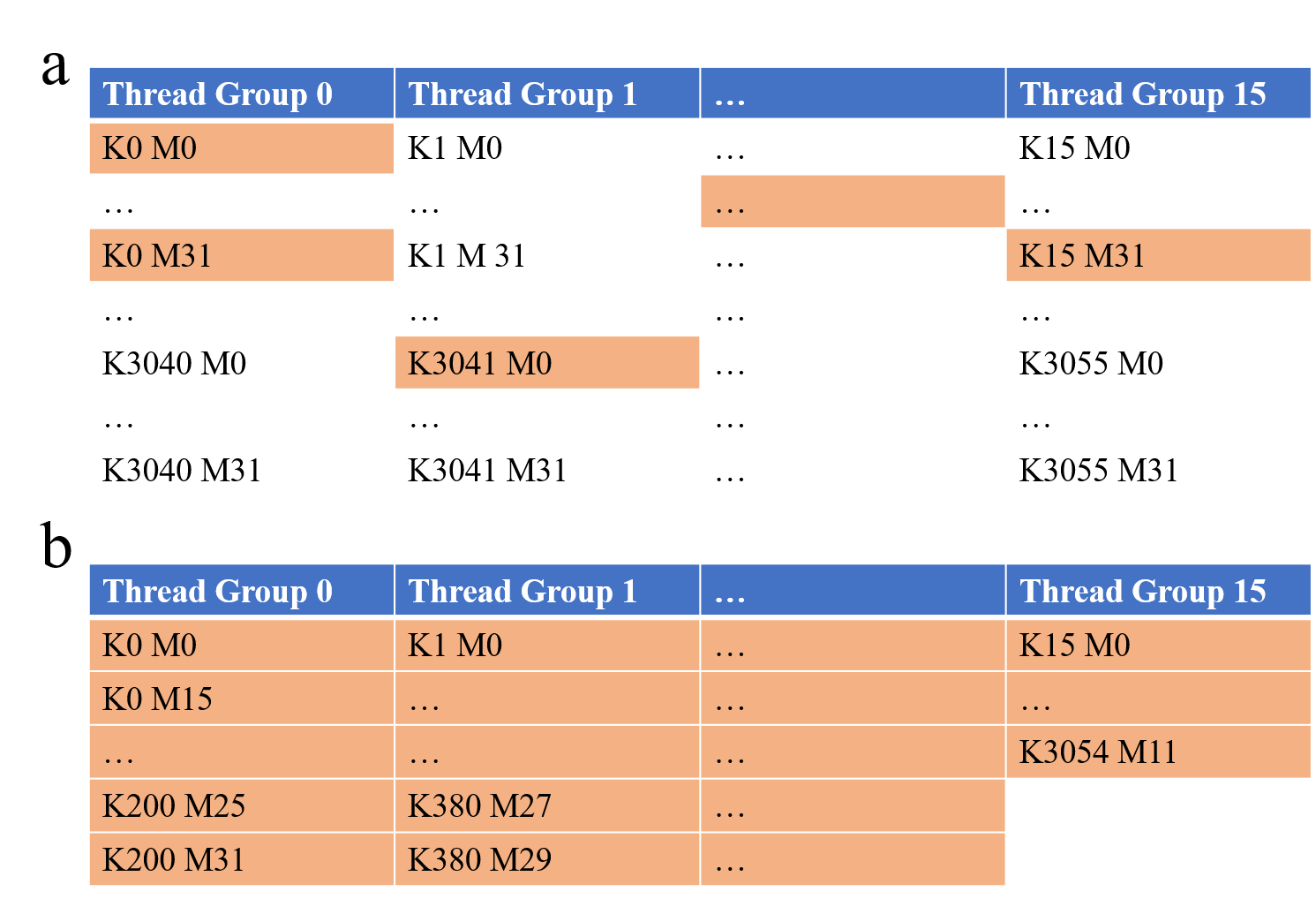}
\end{center}
   \caption{Thread group level load balancing strategies for SpMM ($M=512,K=3056,N=256$). Orange cells denote example nonzero locations. a) The computation that needs to be performed across the different thread groups. b) How the computation is assigned to each thread group.}
\label{fig:groupbalance}
\end{figure}

After we perform the load balancing among thread blocks, we perform the load balancing among the thread groups in the thread block. The computation that needs to be done in a thread block is illustrated in Figure \ref{fig:groupbalance}a. It is based on Figure \ref{fig:tiling}b, but we only need to process the cells in the table corresponding to nonzero elements of the $A$ matrix (illustrated in red). In the dense tiling strategy, each group processes the same number of elements in the reduction axis. In the sparse case, we assign different thread groups different number of elements in the reduction axis so that each thread group except the last one processes the same number of non-zeros, as illustrated in Figure \ref{fig:groupbalance}b. 

Since we know the sparse matrix at compile time, all of the aforementioned load balancing is performed at compile time and used as an input to the code generation.

\subsection{Code Generation}

Each thread group now follows the same basic computation pattern for each element of the reduction axis, as described in Algorithm \ref{alg:spmm}. The thread group loads one element of K in B, $B[b,N_{list}]$ and processes it with nonzero elements in $A[M_{list},b]$. 

To convert the GeMM code to SpMM we use the ``compact'' transformation, first proposed in \cite{venkat2015loop}. We keep only the computations that correspond to nonzero matrix elements in $A$, as illustrated in Figure \ref{fig:groupbalance}a. This corresponds to removing some of the iterations in the loop in lines 9-11 in Algorithm 2. The result is illustrated in Algorithm 3. We propose a novel implementation of the ``compact'' transformation. Typically, the iteration indices corresponding to nonzero elements, $M_{nnz}$, are stored in a separate array, and the code iterates over this data structure, similar to traversing the columns in a CSR data structure. However, this results in dynamic addressing of the accumulators, $ACC$. This results in the usage of local memory instead of registers for $ACC$, which is highly inefficient. 

In SparseRT, we unroll the loop from line 9-11 in the code generator for each thread group and fill in the corresponding sparse matrix values, $A[a,b]$, which we know them at compile time. This leads to the usage of the constant cache for $A$ values and usage of registers for the accumulators. Since the value of the sparse matrix $A$ is broadcast across the thread group, it is an ideal use case for the constant cache. This novel implementation of ``compact'' is critical in achieving good performance. The SpMM code generator is implemented in Python, and generates PTX code.

\begin{algorithm}[tb]
   \caption{SpMM for thread group}
   \label{alg:spmm}
\begin{algorithmic}[1]
   \STATE {\bfseries Input:} matrix pointers $A,B,C$
   \STATE $K_{list}$ = list of B elements to process for thread group
   \STATE $M_{list}$ = list of A elements to process for thread group
   \STATE $N_{list}$ = list of C elements to process for thread group
   \STATE $ACC[M_{list},N_{list}] = {0.0}$ (initialize accumulation buffer)

   \FOR{$b$ in $K_{list}$}
   \STATE $M_{nnz}$ = M indices of nonzero elements in $A[M_{list},b]$
   \STATE Cache $B[b,N_{list}]$ in registers
   \FOR{$a$ in $M_{nnz}$}
    \STATE $ACC[a,N_{list}] += A[a,b] \times B[b,N_{list}]$
   \ENDFOR
   \ENDFOR
\end{algorithmic}
\end{algorithm}

\subsection{Supporting Convolutions}

The SpMM approach thus described can directly handle fully connected layers and 1x1 convolutions. To support convolutions with larger filter sizes, such as the 3x3 convolutions used in ResNet \cite{he2016deep}, we use the im2col transformation to cast the problem into a SpMM. While we refer the reader to \cite{chetlur2014cudnn} for an in-depth description of im2col, the convolution is basically represented as a matrix multiplication by replicating the input activations. Each receptive field for the filter is materialized as a column of the $B$ matrix, whereas each filter is materialized as a row of the $A$ matrix. If the filter weights are pruned, then the $A$ matrix is sparse. 

In order to be efficient, the im2col transformation of the input activations into the dense $B$ matrix needs to be fused into the matrix multiplication kernel. For dense convolutions, it is typically materialized on the fly tile-by-tile as the computation proceeds \cite{chetlur2014cudnn}. For sparse convolutions, $B$ can be treated as a ``virtual'' dense matrix, which is materialized row by row or column by column depending on the format \cite{park2016faster}. 

In SparseRT, we adopt the ``virtual'' dense matrix approach. In the SpMM code, each reference to an element in the $B$ matrix is replaced with the corresponding element in the original input activation tensor. 
The input activations of 3x3 convolutions are typically padded by 1, such that the output of the convolution has the same dimensions as the input. Handling of the padding could require complicated assembly level tricks at the SASS and PTX levels as implemented in cuDNN and ISAAC \cite{chetlur2014cudnn,tillet2017input}. However, in our case, since SparseRT manually unrolls all of the loops in our code, the bounds check required for the padding can be simplified for most input activation locations at compile time. For example, when processing a pixel in the middle of the image, the bounds check can be removed. Once this simplification is done, we notice practically no impact of the padding bounds check on performance. Similar to SpMM, SparseRT directly generates PTX code for sparse convolutions.

\begin{table}
\centering
\caption{Results for SpMM problems in deep learning. Problems are taken from publicly available state-of-the-art pruned neural networks \cite{gale2019state,elsen2019fast} (RN50 = ResNet-50, NLP = Transformer, MbNetV1 = MobileNet V1). Count represents the number of times this SpMM problem dimension appear in the network. Benchmarked on the Tesla T4.}
\begin{tabular}{p{0.8cm}lllp{1.0cm}p{0.6cm}p{0.5cm}p{0.9cm}}
\toprule 

Problem Number & \textit{M} & \textit{K} & \textit{N} & Use Case & Count & \% nnz & Speedup wrt cuBLAS  \\ \midrule
1 & 64 & 256 & 3136 & RN50 & 2 & 0.90 & 4.3 \\ \hline
2 & 256 & 64 & 3136 & RN50 & 2 & 0.90& 2.5  \\ \hline
3 & 128 & 512 & 784 & RN50 & 3 & 0.90  & 4.0  \\ \hline
4 & 512 & 128 & 784 & RN50 & 3 & 0.90 & 3.7 \\ \hline
5 & 256 & 1024 & 196 & RN50 & 5 & 0.90  & 3.0 \\ \hline
6 & 1024 & 256 & 196 & RN50 & 5 & 0.90 & 3.1 \\ \hline
7 & 512 & 2048 & 49 & RN50 & 3 & 0.90 & 2.7 \\ \hline
8 & 2048 & 512 & 49 & RN50 & 3 & 0.90 & 4.8  \\ \hline
1 & 64 & 256 & 3136 & RN50 & 2 & 0.95 & 5.7 \\ \hline
2 & 256 & 64 & 3136 & RN50 & 2 & 0.95&  3.0 \\ \hline
3 & 128 & 512 & 784 & RN50 & 3 & 0.95 & 5.9  \\ \hline
4 & 512 & 128 & 784 & RN50 & 3 & 0.95  & 4.5  \\ \hline
5 & 256 & 1024 & 196 & RN50 & 5 & 0.95 & 4.4 \\ \hline
6 & 1024 & 256 & 196 & RN50 & 5 & 0.95 & 4.2 \\ \hline
7 & 512 & 2048 & 49 & RN50 & 3 & 0.95 & 4.5  \\ \hline
8 & 2048 & 512 & 49 & RN50 & 3 & 0.95  & 6.6 \\ \hline
9 & 2048 & 512 & 256 & NLP & 12 & 0.90 & 4.2  \\ \hline
10 & 512 & 2048 & 256 & NLP & 12 & 0.90  & 4.2  \\ \hline
11 & 512 & 512 & 256 & NLP & 72 & 0.90 & 6.0 \\ \hline
9 & 2048 & 512 & 256 & NLP & 12 & 0.95 & 6.8 \\ \hline
10 & 512 & 2048 & 256 & NLP & 12 & 0.95 & 7.6  \\ \hline
11 & 512 & 512 & 256 & NLP & 72 & 0.95  & 9.0  \\ \hline
12 & 64 & 32 & 12544 & MbNetV1 & 1 & 0.90 & 1.4  \\ \hline
13 & 128 & 64 & 3136 & MbNetV1 & 1 & 0.90 & 3.4 \\ \hline
14 & 128 & 128 & 3136 & MbNetV1 & 1 & 0.90 & 4.0  \\ \hline
15 & 256 & 128 & 784 & MbNetV1 & 1 & 0.90  & 4.2  \\ \hline
16 & 256 & 256 & 784 & MbNetV1 & 1 & 0.90 & 3.5  \\ \hline
17 & 512 & 256 & 196 & MbNetV1 & 1 & 0.90 & 2.3  \\ \hline
18 & 512 & 512 & 196 & MbNetV1 & 5 & 0.90  & 3.2 \\ \hline
19 & 1024 & 512 & 49 & MbNetV1 & 1 & 0.90  & 2.6 \\ \hline
20 & 1024 & 1024 & 49 & MbNetV1 & 1 & 0.90  & 4.5 \\ \hline
\label{tab:results}
\end{tabular}
\end{table}

\subsection{Autotuning}

As aforementioned, there are several parameters in the kernel generation process, $M\_blocks,N\_blocks$ and $Gy$, that can be tuned. We autotune these parameters for each SpMM or sparse convolution problem. Since there are only three such parameters, we perform exhaustive grid search over a list of hand picked choices for each SpMM problem according to a couple of simple heuristics: 1) the split factors, $M\_blocks$, $N\_blocks$ and $Gy$ should be divisible by their respective axes. 2) the block size and the shared memory usage of the kernel is supported by the GPU architecture. This typically resulted in less than 100 parameter combinations for each problem. In practice, the autotuning took less than two minutes for most SpMM and sparse convolution problems. The autotuning was fast in large part because SparseRT directly generates PTX, circumventing Nvidia's NVVM CUDA to PTX compiler. Due to SparseRT's aggressive unrolling strategy, the source code can be large for big sparse matrices, leading to long compilation times with NVVM. This is one of the main motivations for directly generating PTX instead of CUDA.

Theoretically, the best parameters not only depend on the matrix dimensions and sparsity level, but also on the sparsity pattern. In the context of sparse neural network inference, there could be different \textit{instances} of the same SpMM or sparse convolution problem with the same matrix dimensions and sparsity level but different sparsity patterns, since modern neural networks commonly reuse layers. For example, the SpMM operations to be performed for all the 3x3 convolution layers in the first bottleneck group have the same dimensions, yet the sparsity patterns are different. There are up to 72 different instances of SpMM with $M = 512, K = 512, N = 256$ in a Transformer architecture \cite{gale2019state,vaswani2017attention}. If resources allow, optimal performance could be achieved by tuning for each problem instance. However, we notice that in practice, the sparsity patterns for these different instances are similar enough that autotuning on just one of those instances is sufficient.

\begin{figure*}[t]
\begin{center}
\includegraphics[width=\textwidth]{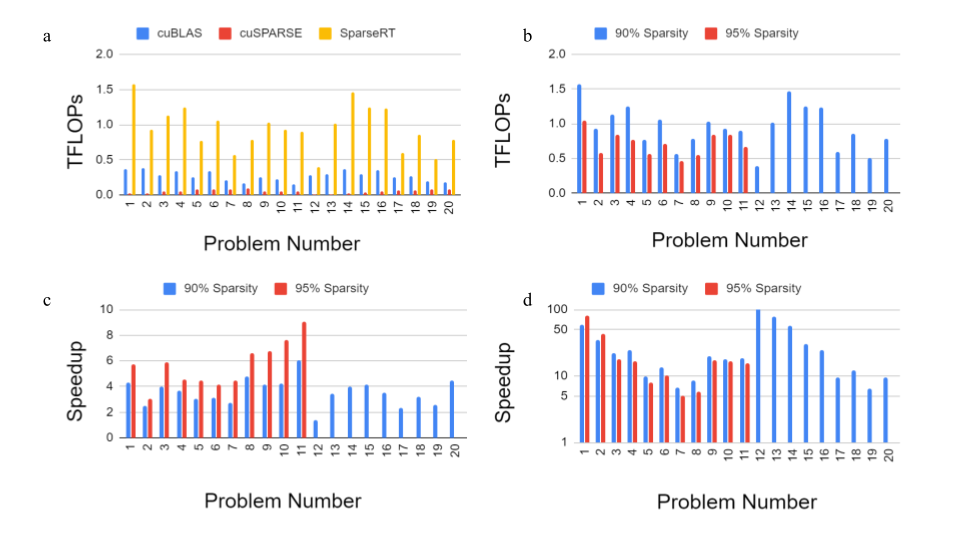}
\end{center}
   \caption{Comparison of our kernels against Nvidia vendor libraries. a) Teraflops per second (TFLOPs) achieved by cuBLAS vs cuSPARSE and SparseRT. cuBLAS performs the sparse matrix multiplication as a dense matrix multiplication, thus requiring 10 times more total FLOPs. We evaluate at 90\% sparsity. b) TFLOPs achieved by SparseRT at different sparsity levels. c) Fold speedup of SparseRT over equivalent dense computation performed by cuBLAS. d) Fold speedup of SparseRT over cuSPARSE. }
\label{fig:flops}
\end{figure*}

\section{Evaluation}

We first evaluate SparseRT's performance on SpMM, then on sparse convolutions. We benchmark our results against three highly optimized vendor libraries, cuDNN, cuBLAS and cuSPARSE. cuBLAS and cuDNN optimize directly at the assembly level and represent the state of the art in dense linear algebra computations on GPU. We use cuDNN v7.6 to benchmark dense 3x3 convolution performance. cuDNN offers different choices for algorithms (FFT vs Winograd vs implicit GeMM). We pick the best performing one for each test case. We use cuBLAS 10.0 to benchmark dense matrix multiplication and 1x1 convolution performance. (We discover that the 1x1 convolution in cuDNN is never faster than the equivalent dense matrix multiplication using cuBLAS.) Similar to cuDNN, cuBLAS offers a variety of algorithms. We try all the available algorithms and select the best performing one for each test case. We use cuSPARSE to benchmark SpMM performance. For cuSPARSE, we use the generic sparse matrix API introduced in 10.1. We use the best-performing sparse matrix format (COO vs CSR) and algorithm for each problem. 

Importantly, while cuSPARSE might appear to be the method of choice in evaluating sparse problems, our empirical observations confirm common knowledge that it is ill-suited for the problem shapes and sparsity levels encounted in deep learning workflows. For this reason, we include dense baselines as the main point of comparison, similar to \cite{gale2020sparse}.

For cuBLAS and cuSPARSE, we try both cases where the weight matrix is transposed or non-transposed. The results for cuBLAS were not significantly different and we report the transposed version to be consistent with our kernel. For cuSPARSE we report the results of the better performing mode. 

We experiment with two recent Nvidia GPU architectures. We benchmark our entire test suite on a Tesla T4 Turing GPU with up to 8.1 TFLOPs of theoretical peak single precision performance. We also benchmark a portion of our test suite on a V100-SXM2 Volta GPU with 15 TFLOPs peak single precision performance to compare performance across different GPU architectures. We use the ``arch=sm\_70'' compilation flag for the V100 and the ``arch=sm\_75'' compilation flag for the T4. The highest (default) optimization option is used with Nvidia's ptxas to compile PTX code. 

\subsection{SpMM}

We evaluate the performance of SparseRT in SpMM on a suite of test problem sizes seen in real-world applications in computer vision (ResNet-50 and MobileNet V1) and natural language processing (Transformer) inference. For ResNet-50 and MobileNet V1 we interpret the 1$\times$1 convolution as a matrix multiplication. We assume a batch size of 1. For the sparse matrix, we use pruned weights from real neural networks. The ResNet-50 and Transformer pruned weights are taken from \cite{gale2019state}, at 90\% and 95\% sparsity levels. The MobileNet weights are taken from \cite{elsen2019fast} at 90\% sparsity level. In Table \ref{tab:acc} we note the relative performance of the pruned models vs. the original dense models, with data from \cite{elsen2019fast} and \cite{gale2019state}. These models represent the best performance achievable at these sparsity levels with three different state-of-the-art pruning methods \cite{gale2019state}. We see that the accuracy loss at 90\% is very limited while the networks suffer a larger accuracy degradation at 95\%. In practice, it is up to practitioners to trade-off between model size, speed and accuracy according to their application requirements. 

\begin{table}
\centering
\caption{Comparison of the relative accuracy achieved by the pruned network vs. the original dense network. Image classification networks are evaluated on top-1 accuracy of ImageNet classification whereas Transformer is evaluated on WMT English-to-German 2014 dataset BLEU score.}
\begin{tabular}{p{2.5cm}p{1cm}p{1.5cm}p{1.5cm}}
\toprule
Network & Sparsity Level & Original Performance & Pruned Performance \\ \hline
ResNet-50 & 90\% & 76.7\% & 75.2\% \\ \hline
ResNet-50 & 95\% & 76.7\% & 72.7\% \\ \hline
Transformer & 90\% & 27.3 & 23.3 \\ \hline
Transformer & 95\% & 27.3 & 20.7 \\ \hline
MobileNet V1 & 90\% & 70.9\% & 68.4\% \\ 
\midrule

\end{tabular}
\label{tab:acc}
\end{table}

This resulted in twenty different SpMM problems, with a total of 257 instances, as listed in Table \ref{tab:results}. The pruned weights are in single precision, which we use for all our experiments. Problem 1-11 were evaluated at two different sparsity levels, 90\% and 95\%. Problems 12-20 were only evaluated at 90\% sparsity because no MobileNet V1 model pruned to 95\% was available. We note that for each SpMM problem, the standard deviation of runtimes across all the different instances is negligible compared to the average runtime, for both SparseRT and cuSPARSE. This indicates that in practice, the different sparsity patterns of the pruned weight matrices do not cause significant variations in performance. The comparison results are presented in Figure \ref{fig:flops}. We make several observations regarding the results. 

\begin{figure*}[t]
\begin{center}
\includegraphics[width=\textwidth]{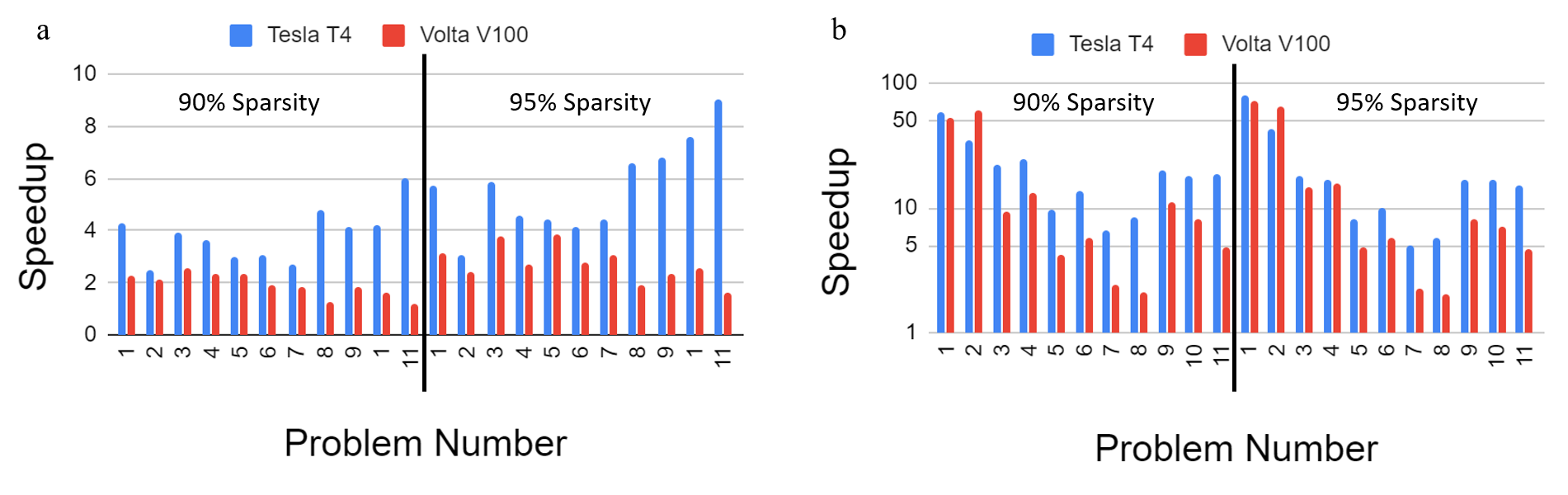}
\end{center}
   \caption{Comparison of performance of SparseRT on two different GPU architectures. Left: Fold speedup of SparseRT over equivalent dense computation by cuBLAS on T4 vs V100. Right: Fold speedup of SparseRT over cuSPARSE on T4 vs V100. }
\label{fig:arch}
\end{figure*}

In Figure \ref{fig:flops}a, we compare the TFLOPs achieved by cuBLAS, cuSPARSE and SparseRT for all the problems at 90\% sparsity. The relative performance at 95\% sparsity is similar. We calculate the TFLOPs by dividing how many total floating point operations are required to perform the SpMM by the time it took each library to execute the operation. For cuBLAS, we perform the equivalent GeMM by representing the sparse matrix as a dense matrix. This results in 90\% wasted work at this sparsity level. Nevertheless, we see that for all test cases, cuBLAS is still more efficient than cuSPARSE. 

Figure \ref{fig:flops}a shows that SparseRT is much more efficient than both vendor libraries for all the problems tested. We notice that the performance of our kernels slightly degrades as $N$ decreases and $M$ and $K$ increase, whereas cuSPARSE's efficiency slightly increases. This finding is consistent with the observation that current SpMM engineering efforts on the GPU, as implemented in cuSPARSE, are focused on cases where the dense matrix is very slim, resembling a collection of vectors. When $N$ decreases or is not a multiple of a power of 2, such as 49 in problem number 7 and 8, the tiling strategies for $N$ is severely limited in SparseRT, leading to further performance degradation. However, SparseRT is still much faster than cuSPARSE and cuBLAS for these problems on the Tesla T4. 

In Figure \ref{fig:flops}b, we compare the efficiency of SparseRT at 90\% vs. 95\% sparsity. As aforementioned, weight matrices at 95\% sparsity is not available for problems 12-20. We see that SparseRT is slightly less efficient as the sparsity level increases from 90\% to 95\%. 

Figure \ref{fig:flops}c shows the speedup achieved by SparseRT over cuBLAS at the two different sparsity levels. We see that halving the number of nonzeros in the problems increases the speedup, but in most cases not by 2x due to the decreased efficiency. We achieve a geometric mean of 3.4x speedup at 90\% sparsity and 5.4x speedup at 95\% sparsity. We see that the levels of speedup attainable for different SpMM problems are highly variable, a phenomenon that will be discussed in detail later. 

Figure \ref{fig:flops}d shows the speedup achieved by SparseRT over cuSPARSE (Figure \ref{fig:flops}d). Overall, we see that the speedups are highly consistent between the different sparsity levels. We achieve a geometric mean of 20x speedup at 90\% sparsity on problems 1-20 and 15x speedup at 95\% sparsity on problems 1-11. These results support common observations that current state-of-the-art vendor libraries for sparse linear algebra such as cuSPARSE are ill-suited for sparse deep learning applications. 

To measure the effect of GPU architecture on SparseRT, we also run SpMM problems 1-11 on a Volta V100 at both sparsity levels. In Figure \ref{fig:arch}, we compare the relative performance of SparseRT on these two recent GPU architectures. In Figure \ref{fig:arch}a, we show the speedup achieved over cuBLAS for both GPUs. We see that SparseRT does better on Tesla T4 in all cases. However the relative performance is consistent -- problems where SparseRT showed larger performance gains on the T4 also showed relatively larger gains on the V100. On the V100, SparseRT is still faster than cuBLAS in all cases. In Figure \ref{fig:arch}b, we compare SparseRT's speedup over cuSPARSE on both GPUs. We see that SparseRT's relative performance is stable across both GPUs, since cuSPARSE is also slower relative to cuBLAS on the V100. SparseRT is faster than cuSPARSE in all cases. We conclude that SparseRT is portable and robust across different GPU generations. 

\begin{figure}
\begin{center}
\includegraphics[width=\linewidth]{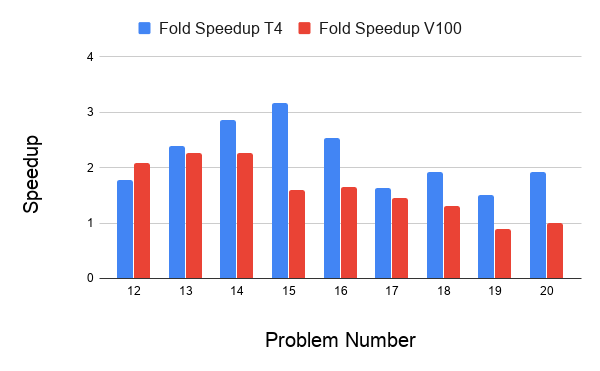}
\end{center}
   \caption{Performance of SparseRT vs Sputnik on the SpMM problems in MobileNet V1.}
\label{fig:sputnik}
\end{figure}

To further study the performance of SparseRT, we compare it with Sputnik, a recently released sparse linear algebra library for deep neural networks on GPUs\cite{gale2020sparse} . Sputnik uses a combination of optimization techniques such as row reordering, memory alignment and subwarp tiling. It offers a huge improvement over state-of-the-art baselines such as ASpT \cite{hong2019adaptive} and cuSPARSE, demonstrating end to end speedups in inference with MobileNet V1. We compare SparseRT's performance with Sputnik on the SpMM problems encountered in MobileNet V1. In Figure \ref{fig:sputnik} we see that SparseRT outperforms Sputnik on problems with smaller sparse matrices and achieves comparable performance on problems with larger sparse matrices. We will compare SparseRT's approach and Sputnik's approach in detail in Section 5.

\subsection{Sparse Convolution}

\begin{figure}
\begin{center}
\includegraphics[width=\linewidth]{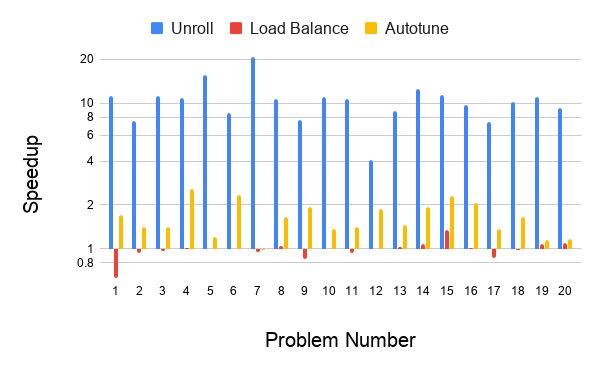}
\end{center}
   \caption{Performance of SparseRT's sparse convolution vs dense convolution by cuDNN at two different sparsity levels. All convolutions have 3x3 filters and padding size of 1.}
\label{fig:conv}
\end{figure}

We benchmark the performance of SparseRT on four 3x3 convolutions use cases in ResNet-50 on the T4. We assume a batch size of 1. The pruned weights come from the same model used to evaluate the 1x1 convolutions \cite{gale2019state}. We evaluate the sparse convolution problems at two different sparsity levels, as shown in Table \ref{tab:3xconv} and Figure \ref{fig:conv}. The filter size used for all the problems are 3 by 3. The padding size is 1. We obtain significant speedups at both sparsity levels for all three problems. Similar to SpMM, we find that different instances of sparse convolution problems with the same dimensions, corresponding to different layers in the same block group in ResNet-50, have virtually the same runtime. 

As expected, the cuDNN baseline does not use the im2col + GeMM approach for all the problems tested. Instead, the Winograd algorithm is used, with im2col + GeMM being more than 2x slower. This means the speedup of SparseRT's im2col + SpMM convolution over the comparable im2col + GeMM convolution in cuDNN is even larger than the numbers presented in Table \ref{tab:3xconv}, confirming the efficiency of our SpMM implementation.

Similarly to SpMM, we see that as the sparse weights size increase, the speedups obtained decreases, with the notable exception of problem number 4. We will describe the cause in the next section.

\begin{table}
\centering
\caption{Sparse convolution problems in deep learning. All filter sizes are 3x3 with a padding of 1. Problems are taken from pruned ResNet-50 \cite{gale2019state}.}
\begin{tabular}{p{1cm}p{1cm}p{1cm}p{1cm}p{1cm}p{1cm}}
\toprule
Problem Number & Image Dimension & Input Channels & Output Channels & Sparsity &  Speedup \\ 
\midrule
1 & 56 & 64 & 64 & 0.9 & 3.7 \\ \hline
2 & 28 & 128 & 128 & 0.9 & 2.0 \\ \hline
3 & 14 & 256 & 256 & 0.9 & 1.4 \\ \hline
4 & 7 & 512 & 512 & 0.9 & 2.4 \\ \hline
1 & 56 & 64 & 64 & 0.95 & 5.3 \\ \hline
2 & 28 & 128 & 128 & 0.95 & 3.5 \\ \hline
3 & 14 & 256 & 256 & 0.95 & 2.5 \\ \hline
4 & 7 & 512 & 512 & 0.95 & 8.5 \\ \hline

\end{tabular}
\label{tab:3xconv}
\end{table}

\subsection{Ablation Studies and Performance Analysis}
From Figure \ref{fig:flops}, we can see that SparseRT achieves the most speedup over cuBLAS and cuSPARSE when the sparse matrix is relatively small compared to the dense matrix, for example problem number 1 and 3. In cases where the sparse matrix is much larger than the dense matrix, such as problem number 6 or 7, SparseRT cannot offer as much speedup over cuBLAS. How do we understand this phenomenon?

We note that in dense matrix multiplication (GeMM), the asymptotic runtime is proportional to $MNK$. This is because optimized GeMM routines in cuBLAS are typically compute-bound, and the dense GeMM has $O(MNK)$ multiplications. Deep learning architects have thus designed networks keeping this in mind -- in earlier layers, the activation filter map is large ($N$) but the number of channels ($M$ and $K$) are small; in later layers, the situation is reversed. As a result, the GeMM runtime doesn't change much for all the matrix multiplications in a network, as seen in Figure \ref{fig:flops}a.

\begin{figure}
\begin{center}
\includegraphics[width=\linewidth]{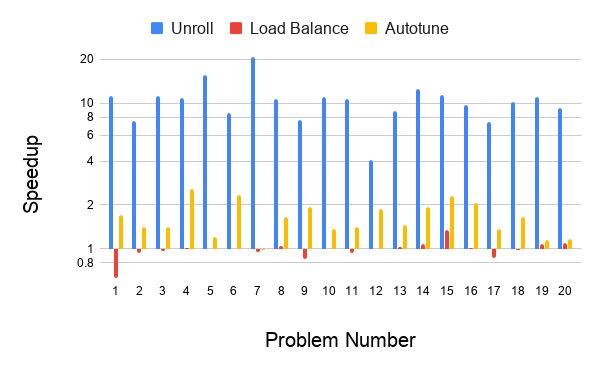}
\end{center}
   \caption{Speedup provided by code unrolling, static load balancing and autotuning in SparseRT, measured on SpMM problems at 90\% sparsity.}
\label{fig:nounroll}
\end{figure}

Sparse kernels, on the other hand, are typically memory-bound instead of compute-bound. SparseRT employs aggressive code unrolling approaches that bake all the sparse matrix information into the source code. Instead of fetching both matrices from the data cache, we are now fetching one from the shared instruction and constants cache and another from the on-chip data cache to minimize cache contention. In Figure \ref{fig:nounroll}, we compare unrolled SparseRT code vs. not unrolled code that uses the same tiling strategy on the SpMM problems at 90\% sparsity. We see that SparseRT's unrolling strategy can lead to over 10x speedups. 

The code unrolling strategy means that for problems with very large sparse matrices, SparseRT kernels are often bottlenecked by instruction fetch latency instead of memory or execution dependency. When $M$ and $K$ increase, the 64Kb constant cache is depleted. The compile time constants are now compiled to immediate constants in move instructions, which is less efficient than using the constant cache. As a result, we see that the SparseRT kernel runtimes are roughly proportional to $MK$, instead of $MNK$. However, SparseRT is still more efficient than cuBLAS in cases where $MK$ is very large, such as problem number 6 or 7. The comparison with Sputnik in Figure \ref{fig:sputnik} confirms this trend: while SparseRT generally outperforms Sputnik in cases where the sparse matrix is small, Sputnik catches up with SparseRT on larger sparse matrix sizes.

cuSPARSE kernel runtimes exhibit the opposite trend as SparseRT kernels. As the dense matrix size relative to the sparse matrix, cuSPARSE does better. This is expected as cuSPARSE is catered to use cases in scientific computing, where the sparse matrix is often massive compared to the dense matrix. We can see in Figure \ref{fig:flops}d that while SparseRT's advantage over cuSPARSE is massive ($>$50x) for SpMM problems in earlier layers of neural networks, where $M$ and $K$ are very small, it quickly degrades as $M$ and $K$ increases relative to $N$.

In this paper, we are evaluating neural networks assuming an input size of 224 by 224. In real use cases, the input can be considerably larger. For example, certain automotive and retail applications require video resolution of more than 1 million pixels. Since increasing the input size increases $N$ instead of $M$ or $K$ (assuming same network architecture), we expect SparseRT to enjoy an even greater advantage over cuBLAS in those cases. 

We also did ablation studies to examine the effects of our static load balancing scheme and autotuning on performance, as shown in Figure \ref{fig:nounroll}. When we do not load balance the sparse matrix across thread blocks, we see comparable performance in most cases. In a few of those problems, we even saw a performance improvement. This could be attributed to the fact that the sparsity patterns of the sparse matrices obtained from common pruning techniques are quite uniform to begin with. The static load balancing techniques used here that allocate different number of sparse matrix rows to different thread blocks uses different numbers of accumulator registers per thread block. This could lead to wasted resources, as the GPU hardware cannot dynamically allocate registers to thread blocks and must assign the maximum number of registers requested by all thread blocks to each thread block. 

To study the effects of autotuning, we fix a single parameter choice for each problem. This resulted in around a 2x slowdown for most of the problems selected, which proves the benefit of autotuning to our approach. Since we are optimizing for inference, the autotuning cost can be effectively amortized over all subsequent inference requests. 

\section{Comparison to Related Work}

In this work, we present an approach to accelerate SpMM computations encountered in deep learning applications. We show that we can achieve a geometric mean of 3.4x speedup at 90\% sparsity and 5.4x speedup at 95\% sparsity over the dense computation for SpMM and similarly for sparse convolution. This level of performance is often an order of magnitude better (up to 60x) than what vendor sparse linear algebra libraries such as cuSPARSE can achieve. It is also much better than state-of-the-art libraries such as Sputnik for some SpMM problems and competitive on others. Notably, current structured pruning approaches can also only achieve up to around 2x - 3x inference speedup on GPU \cite{he2017channel,lin2019toward,yao2019balanced}. 

Our approach shows 2-3x speedups over Sputnik on smaller matrices and comparable performance on larger matrices. This trend can be expected given the relatively poorer performance of SparseRT on larger sparse matrices. SparseRT enjoys a larger advantage on the Tesla T4 than on the V100. This difference could be attributed to the fact that Sputnik was tuned to target the V100 architecture \cite{gale2020sparse}. While SparseRT and Sputnik share some common optimization strategies, such as parallelizing the dense matrix columns across thread blocks, they are fundamentally different approaches. Sputnik stores the sparse matrix data in shared memory, and is bottlenecked by shared memory loads. Therefore, its novel optimizations, subwarp tiling and reverse offset memory alignment, are targeted towards mitigating the load bandwidth bottleneck by enabling vector memory loads. In SparseRT, the sparse matrix data is stored exclusively in the instruction cache baked into instruction operands, so these optimizations are not necessary. On the other hand, Sputnik does not suffer from instruction fetch bottleneck for larger matrices, and its performance does not degrade.

Interestingly, Sputnik also employs a row-swizzling preprocessing strategy on the sparse matrix to improve load balancing. Such a preprocessing strategy could benefit SparseRT as well, albeit for a different reason. Row swizzling puts rows with similar nonzero patterns next to each other, which will lead to a fortuitous reduction of dense matrix memory loads if adjacent sparse matrix rows have nonzeros at the same column positions. In deep learning inference, preprocessing sparse weights can be quite effective as its cost can be amortized over all subsequent inference requests. We are currently exploring preprocessing strategies to improve SparseRT's performance. 

We did not compare SparseRT with ASpT in \cite{hong2019adaptive} and MergeSpMM in \cite{yang2018design} as they were compared with Sputnik in \cite{gale2020sparse} and shown to be inferior on almost all deep learning SpMM problems considered. ASpT and MergeSpMM exemplify optimizations for SpMM workloads in scientific computing where the sparse matrix is large and highly sparse, while the dense matrix is ``tall and skinny'', with a fixed and small number of columns. In fact, the implementations for ASpT and MergeSpMM both expect a dense matrix with either 32 or 128 columns. Unfortunately, in deep learning, the sparse matrix can be smaller than the dense matrix and the sparsity ratio is often much lower than applications encountered in scientific computing. In addition, the dense matrix can have any number of columns. In addition, the sparsity pattern produced by common pruning techniques such as magnitude pruning is usually quite uniform, in contrast to power-law or blocked patterns encountered in scientific computing. This makes our static load balancing strategy not very effective on most sparse matrices, and nullifies the motivation for approaches such as ASpT, which seek to take advantage of different sparsity patterns in different regions of the sparse matrix. However, similar to Sputnik's row-swizzling strategy, interesting preprocessing strategies have been proposed for ASpT that could also benefit SparseRT \cite{jiang2020novel}.

SpMM can be used for sparse 1x1 convolutions in deep learning. Some deep neural networks, such as ResNet-50, also make heavy use of 3x3 convolutions. There has been several recent works on accelerating sparse 3x3 convolutions, such as SkimCaffe and Tiramisu on CPUs and Escort on GPUs \cite{baghdaditiramisu,park2016faster,chen2019escoin}. SparseRT adopts the ``virtual'' dense matrix approach to fuse the im2col transformation, similar to SkimCaffe and Tiramisu \cite{park2016faster,baghdaditiramisu}. Escort opts for a direct sparse convolution approach where the convolution is not cast as an implicit GeMM. Instead, the classic 7-loop convolution algorithm is carried out. Escort is two to three times faster than im2col + cuBLAS GeMM for several sparse convolution problems in deep learning \cite{chen2019escoin}. However, the latest cuDNN library can be several times faster than im2col + cuBLAS GeMM, using optimizations such as the Winograd algorithm. For example, on the four problems listed in this paper, cuDNN's Winograd convolution is anywhere between two to four times faster than cuDNN's implicit GeMM convolution. Thus, we only compare performance of SparseRT with cuDNN, and not with Escort in this work. However, there are parallel ideas between this work and Escort. Both recognize that the problem is memory bound. While Escort aims to alleviate memory bottlenecks through putting the sparse matrix and the dense matrix into different forms of on-chip data caches, SparseRT puts the sparse matrix into the instruction cache and constant cache, avoiding the on-chip data caches altogether. Direct sparse convolution approaches also aim to remove the overhead from the im2col transformation. We show that by adopting the ``virtual'' dense matrix technique from \cite{park2016faster}, we can effectively fuse the im2col transformation with the SpMM and retain good performance. 

\section{Discussion and Conclusion}

Many complementary approaches have been proposed to accelerate neural network inference, such as distillation, quantization and pruning \cite{hinton2015distilling,han2015deep}. This work presents SparseRT, a system to support efficient inference with unstructured weight pruning. While SparseRT currently only considers weight sparsity, weights are not the only source of sparsity in an neural network. Activation functions sometimes produce sparse input activations. Although in some tasks such as LiDAR and face detection, the input might be sparse enough to offer performance improvement \cite{dong2019acorns}, this is not the case for most deep learning tasks. In addition, the input sparsity pattern depends on the specific input example, making static optimizations employed by SparseRT difficult. As a result of the narrow applicability and challenging implementation, SparseRT currently does not support activation sparsity. 

There has also been significant recent effort dedicated to reducing the precision of the data types used in the neural network, especially during inference. However, recent pruning results in literature still rely on floating point \cite{gale2019state}. As a result, in this work, we focus on single precision computations. We are working on extending our approach to support unstructured sparsity with lower numerical precision through leveraging the GPU vector instructions, such as dp4a. We will not be able to leverage dense matrix multiplication accelerators such as tensor cores that support these lower precision data types. This is a fundamental limitation of unstructured sparsity, though recent research may suggest solutions to this problem by augmenting the dense accelerator architectures \cite{zhu2019sparse}. Recent commercial architectures such as the Nvidia A100 are promising developments in this direction.

In addition, it must be noted that the current work only targets the matrix multiplications and convolutions in the neural network. The performance of modern DNNs is often bottlenecked by other operations, such as softmax and self-attention. Currently, SparseRT does not support these operations, because they are typically not done in a sparse manner even in pruned neural networks. For example, in Transformer networks, even though the weight matrix used to generate the query and key vectors can be pruned \cite{gale2019state}, the query and key vectors themselves are still dense, making the self-attention a dense operation. We expect future research in neural network compression and pruning to allow these operators to become sparse as well. 

Although currently applied to sparse inference, SparseRT can also potentially be applied to sparse training. Current sparse training algorithms either use a fixed sparse neural network architecture, or fixes a particular architecture for a number of iterations \cite{evci2019rigging,frankle2018lottery}. This suggests that we can either autotune kernels for the chosen sparse neural net at the onset or just-in-time as the architecture changes. SparseRT currently does not support sparse gradient computations, which are typically sampled dense-dense matrix multiplications. 

Finally, it is important to note again that the current work focuses on the optimization of single layers in the neural network. We refrain from presenting end to end neural network inference timing results here because there could be a variety of confounding factors due to inter-layer optimizations employed by state-of-the-art inference engines such as TensorRT \cite{tensorrt}. Instead, we focus here on demonstrating that key primitives in deep learning inference could be performed efficiently with unstructured sparsity. 

In conclusion, we present SparseRT, an approach to accelerate unstructured sparsity on GPUs for deep learning inference based on the inspector-executor framework \cite{venkat2015loop}. We present significant speedups on a test suite of hundreds of SpMM and sparse convolution problems in deep learning. Future work include supporting other sparse operations such as sampled dense matrix matrix multiplication, lower precision data formats such as int8, and porting to other architectures such as multicore CPUs with vector instructions. We hope this work challenges the longstanding belief amongst practitioners that unstructured sparsity is poorly supported on modern parallel architectures, and inspires further research on unstructured sparse pruning methods.

\section{Acknowledgments}

The author thanks Fredrik Kjolstad, Philippe Tillet and Ajay Brahmakshatriya for helpful discussions and comments. The author thanks Trevor Gale for helping with Sputnik benchmarks.

\bibliographystyle{ACM-Reference-Format}
\bibliography{tile}


\begin{thebibliography}{47}


\ifx \showCODEN    \undefined \def \showCODEN     #1{\unskip}     \fi
\ifx \showDOI      \undefined \def \showDOI       #1{#1}\fi
\ifx \showISBNx    \undefined \def \showISBNx     #1{\unskip}     \fi
\ifx \showISBNxiii \undefined \def \showISBNxiii  #1{\unskip}     \fi
\ifx \showISSN     \undefined \def \showISSN      #1{\unskip}     \fi
\ifx \showLCCN     \undefined \def \showLCCN      #1{\unskip}     \fi
\ifx \shownote     \undefined \def \shownote      #1{#1}          \fi
\ifx \showarticletitle \undefined \def \showarticletitle #1{#1}   \fi
\ifx \showURL      \undefined \def \showURL       {\relax}        \fi
\providecommand\bibfield[2]{#2}
\providecommand\bibinfo[2]{#2}
\providecommand\natexlab[1]{#1}
\providecommand\showeprint[2][]{arXiv:#2}

\bibitem[\protect\citeauthoryear{??}{int}{[n.d.]}]%
        {intelieblas}
 \bibinfo{year}{[n.d.]}\natexlab{}.
\newblock \bibinfo{title}{Intel Inspector Executor Sparse BLAS}.
\newblock
  \bibinfo{howpublished}{\url{https://software.intel.com/en-us/mkl-developer-reference-c-inspector-executor-sparse-blas-routines}}.
\newblock


\bibitem[\protect\citeauthoryear{??}{ten}{[n.d.]}]%
        {tensorrt}
 \bibinfo{year}{[n.d.]}\natexlab{}.
\newblock \bibinfo{title}{TensorRT}.
\newblock \bibinfo{howpublished}{\url{https://developer.nvidia.com/tensorrt}}.
\newblock


\bibitem[\protect\citeauthoryear{Ansel, Kamil, Veeramachaneni, Ragan-Kelley,
  Bosboom, O'Reilly, and Amarasinghe}{Ansel et~al\mbox{.}}{2014}]%
        {ansel2014opentuner}
\bibfield{author}{\bibinfo{person}{Jason Ansel}, \bibinfo{person}{Shoaib
  Kamil}, \bibinfo{person}{Kalyan Veeramachaneni}, \bibinfo{person}{Jonathan
  Ragan-Kelley}, \bibinfo{person}{Jeffrey Bosboom}, \bibinfo{person}{Una-May
  O'Reilly}, {and} \bibinfo{person}{Saman Amarasinghe}.}
  \bibinfo{year}{2014}\natexlab{}.
\newblock \showarticletitle{Opentuner: An extensible framework for program
  autotuning}. In \bibinfo{booktitle}{\emph{Proceedings of the 23rd
  international conference on Parallel architectures and compilation}}. ACM,
  \bibinfo{pages}{303--316}.
\newblock


\bibitem[\protect\citeauthoryear{Baghdadi, Debbagh, Abdous, Zohra, Renda,
  Frankle, Carbin, and Amarasinghe}{Baghdadi et~al\mbox{.}}{[n.d.]}]%
        {baghdaditiramisu}
\bibfield{author}{\bibinfo{person}{Riyadh Baghdadi},
  \bibinfo{person}{Abdelkader~Nadir Debbagh}, \bibinfo{person}{Kamel Abdous},
  \bibinfo{person}{Benhamida~Fatima Zohra}, \bibinfo{person}{Alex Renda},
  \bibinfo{person}{Jonathan~Elliott Frankle}, \bibinfo{person}{Michael Carbin},
  {and} \bibinfo{person}{Saman Amarasinghe}.}
  \bibinfo{year}{[n.d.]}\natexlab{}.
\newblock \showarticletitle{TIRAMISU: A Polyhedral Compiler for Dense and
  Sparse Deep Learning}.
\newblock  (\bibinfo{year}{[n.\,d.]}).
\newblock


\bibitem[\protect\citeauthoryear{Chen, Moreau, Jiang, Shen, Yan, Wang, Hu,
  Ceze, Guestrin, and Krishnamurthy}{Chen et~al\mbox{.}}{2018}]%
        {chen2018tvm}
\bibfield{author}{\bibinfo{person}{Tianqi Chen}, \bibinfo{person}{Thierry
  Moreau}, \bibinfo{person}{Ziheng Jiang}, \bibinfo{person}{Haichen Shen},
  \bibinfo{person}{Eddie~Q Yan}, \bibinfo{person}{Leyuan Wang},
  \bibinfo{person}{Yuwei Hu}, \bibinfo{person}{Luis Ceze},
  \bibinfo{person}{Carlos Guestrin}, {and} \bibinfo{person}{Arvind
  Krishnamurthy}.} \bibinfo{year}{2018}\natexlab{}.
\newblock \showarticletitle{TVM: end-to-end optimization stack for deep
  learning}.
\newblock \bibinfo{journal}{\emph{arXiv preprint arXiv:1802.04799}}
  (\bibinfo{year}{2018}), \bibinfo{pages}{1--15}.
\newblock


\bibitem[\protect\citeauthoryear{Chen}{Chen}{2019}]%
        {chen2019escoin}
\bibfield{author}{\bibinfo{person}{Xuhao Chen}.}
  \bibinfo{year}{2019}\natexlab{}.
\newblock \showarticletitle{Escoin: Efficient Sparse Convolutional Neural
  Network Inference on GPUs}.
\newblock \bibinfo{journal}{\emph{arXiv}} (\bibinfo{year}{2019}).
\newblock


\bibitem[\protect\citeauthoryear{Chetlur, Woolley, Vandermersch, Cohen, Tran,
  Catanzaro, and Shelhamer}{Chetlur et~al\mbox{.}}{2014}]%
        {chetlur2014cudnn}
\bibfield{author}{\bibinfo{person}{Sharan Chetlur}, \bibinfo{person}{Cliff
  Woolley}, \bibinfo{person}{Philippe Vandermersch}, \bibinfo{person}{Jonathan
  Cohen}, \bibinfo{person}{John Tran}, \bibinfo{person}{Bryan Catanzaro}, {and}
  \bibinfo{person}{Evan Shelhamer}.} \bibinfo{year}{2014}\natexlab{}.
\newblock \showarticletitle{cudnn: Efficient primitives for deep learning}.
\newblock \bibinfo{journal}{\emph{arXiv preprint arXiv:1410.0759}}
  (\bibinfo{year}{2014}).
\newblock


\bibitem[\protect\citeauthoryear{Crowley, Turner, Storkey, and Michael}{Crowley
  et~al\mbox{.}}{2019}]%
        {closerlook2019}
\bibfield{author}{\bibinfo{person}{Elliot Crowley}, \bibinfo{person}{Jack
  Turner}, \bibinfo{person}{Amos Storkey}, {and} \bibinfo{person}{O'Boyle
  Michael}.} \bibinfo{year}{2019}\natexlab{}.
\newblock \showarticletitle{A closer look at structured pruning for neural
  network compression}.
\newblock \bibinfo{journal}{\emph{arXiv preprint arXiv:1810.04622v3}}
  (\bibinfo{year}{2019}).
\newblock


\bibitem[\protect\citeauthoryear{Devlin, Chang, Lee, and Toutanova}{Devlin
  et~al\mbox{.}}{2018}]%
        {devlin2018bert}
\bibfield{author}{\bibinfo{person}{Jacob Devlin}, \bibinfo{person}{Ming-Wei
  Chang}, \bibinfo{person}{Kenton Lee}, {and} \bibinfo{person}{Kristina
  Toutanova}.} \bibinfo{year}{2018}\natexlab{}.
\newblock \showarticletitle{Bert: Pre-training of deep bidirectional
  transformers for language understanding}.
\newblock \bibinfo{journal}{\emph{arXiv preprint arXiv:1810.04805}}
  (\bibinfo{year}{2018}).
\newblock


\bibitem[\protect\citeauthoryear{Dong, Liu, Zhao, Li, Li, Wang, and Feng}{Dong
  et~al\mbox{.}}{2019}]%
        {dong2019acorns}
\bibfield{author}{\bibinfo{person}{Xiao Dong}, \bibinfo{person}{Lei Liu},
  \bibinfo{person}{Peng Zhao}, \bibinfo{person}{Guangli Li},
  \bibinfo{person}{Jiansong Li}, \bibinfo{person}{Xueying Wang}, {and}
  \bibinfo{person}{Xiaobing Feng}.} \bibinfo{year}{2019}\natexlab{}.
\newblock \showarticletitle{Acorns: A framework for accelerating deep neural
  networks with input sparsity}. In \bibinfo{booktitle}{\emph{2019 28th
  International Conference on Parallel Architectures and Compilation Techniques
  (PACT)}}. IEEE, \bibinfo{pages}{178--191}.
\newblock


\bibitem[\protect\citeauthoryear{Elsen, Dukhan, Gale, and Simonyan}{Elsen
  et~al\mbox{.}}{2019}]%
        {elsen2019fast}
\bibfield{author}{\bibinfo{person}{Erich Elsen}, \bibinfo{person}{Marat
  Dukhan}, \bibinfo{person}{Trevor Gale}, {and} \bibinfo{person}{Karen
  Simonyan}.} \bibinfo{year}{2019}\natexlab{}.
\newblock \showarticletitle{Fast Sparse ConvNets}.
\newblock \bibinfo{journal}{\emph{arXiv preprint arXiv:1911.09723}}
  (\bibinfo{year}{2019}).
\newblock


\bibitem[\protect\citeauthoryear{Evci, Gale, Menick, Castro, and Elsen}{Evci
  et~al\mbox{.}}{2019}]%
        {evci2019rigging}
\bibfield{author}{\bibinfo{person}{Utku Evci}, \bibinfo{person}{Trevor Gale},
  \bibinfo{person}{Jacob Menick}, \bibinfo{person}{Pablo~Samuel Castro}, {and}
  \bibinfo{person}{Erich Elsen}.} \bibinfo{year}{2019}\natexlab{}.
\newblock \showarticletitle{Rigging the Lottery: Making All Tickets Winners}.
\newblock \bibinfo{journal}{\emph{arXiv preprint arXiv:1911.11134}}
  (\bibinfo{year}{2019}).
\newblock


\bibitem[\protect\citeauthoryear{Frankle and Carbin}{Frankle and
  Carbin}{2018}]%
        {frankle2018lottery}
\bibfield{author}{\bibinfo{person}{Jonathan Frankle} {and}
  \bibinfo{person}{Michael Carbin}.} \bibinfo{year}{2018}\natexlab{}.
\newblock \showarticletitle{The lottery ticket hypothesis: Finding sparse,
  trainable neural networks}.
\newblock \bibinfo{journal}{\emph{arXiv preprint arXiv:1803.03635}}
  (\bibinfo{year}{2018}).
\newblock


\bibitem[\protect\citeauthoryear{Gale, Elsen, and Hooker}{Gale
  et~al\mbox{.}}{2019}]%
        {gale2019state}
\bibfield{author}{\bibinfo{person}{Trevor Gale}, \bibinfo{person}{Erich Elsen},
  {and} \bibinfo{person}{Sara Hooker}.} \bibinfo{year}{2019}\natexlab{}.
\newblock \showarticletitle{The state of sparsity in deep neural networks}.
\newblock \bibinfo{journal}{\emph{arXiv preprint arXiv:1902.09574}}
  (\bibinfo{year}{2019}).
\newblock


\bibitem[\protect\citeauthoryear{Gale, Zaharia, Young, and Elsen}{Gale
  et~al\mbox{.}}{2020}]%
        {gale2020sparse}
\bibfield{author}{\bibinfo{person}{Trevor Gale}, \bibinfo{person}{Matei
  Zaharia}, \bibinfo{person}{Cliff Young}, {and} \bibinfo{person}{Erich
  Elsen}.} \bibinfo{year}{2020}\natexlab{}.
\newblock \showarticletitle{Sparse GPU Kernels for Deep Learning}.
\newblock \bibinfo{journal}{\emph{arXiv preprint arXiv:2006.10901}}
  (\bibinfo{year}{2020}).
\newblock


\bibitem[\protect\citeauthoryear{Gray, Radford, and Kingma}{Gray
  et~al\mbox{.}}{2017}]%
        {gray2017gpu}
\bibfield{author}{\bibinfo{person}{Scott Gray}, \bibinfo{person}{Alec Radford},
  {and} \bibinfo{person}{Diederik~P Kingma}.} \bibinfo{year}{2017}\natexlab{}.
\newblock \showarticletitle{Gpu kernels for block-sparse weights}.
\newblock \bibinfo{journal}{\emph{arXiv preprint arXiv:1711.09224}}
  (\bibinfo{year}{2017}).
\newblock


\bibitem[\protect\citeauthoryear{Han, Liu, Mao, Pu, Pedram, Horowitz, and
  Dally}{Han et~al\mbox{.}}{2016}]%
        {han2016eie}
\bibfield{author}{\bibinfo{person}{Song Han}, \bibinfo{person}{Xingyu Liu},
  \bibinfo{person}{Huizi Mao}, \bibinfo{person}{Jing Pu},
  \bibinfo{person}{Ardavan Pedram}, \bibinfo{person}{Mark~A Horowitz}, {and}
  \bibinfo{person}{William~J Dally}.} \bibinfo{year}{2016}\natexlab{}.
\newblock \showarticletitle{EIE: efficient inference engine on compressed deep
  neural network}. In \bibinfo{booktitle}{\emph{2016 ACM/IEEE 43rd Annual
  International Symposium on Computer Architecture (ISCA)}}. IEEE,
  \bibinfo{pages}{243--254}.
\newblock


\bibitem[\protect\citeauthoryear{Han, Mao, and Dally}{Han
  et~al\mbox{.}}{2015}]%
        {han2015deep}
\bibfield{author}{\bibinfo{person}{Song Han}, \bibinfo{person}{Huizi Mao},
  {and} \bibinfo{person}{William~J Dally}.} \bibinfo{year}{2015}\natexlab{}.
\newblock \showarticletitle{Deep compression: Compressing deep neural networks
  with pruning, trained quantization and huffman coding}.
\newblock \bibinfo{journal}{\emph{arXiv preprint arXiv:1510.00149}}
  (\bibinfo{year}{2015}).
\newblock


\bibitem[\protect\citeauthoryear{Harris and Perelygin}{Harris and
  Perelygin}{2017}]%
        {harris2017cooperative}
\bibfield{author}{\bibinfo{person}{M Harris} {and} \bibinfo{person}{K
  Perelygin}.} \bibinfo{year}{2017}\natexlab{}.
\newblock \bibinfo{title}{Cooperative groups: Flexible CUDA thread
  programming}.
\newblock
\newblock


\bibitem[\protect\citeauthoryear{He, Zhang, Ren, and Sun}{He
  et~al\mbox{.}}{2016}]%
        {he2016deep}
\bibfield{author}{\bibinfo{person}{Kaiming He}, \bibinfo{person}{Xiangyu
  Zhang}, \bibinfo{person}{Shaoqing Ren}, {and} \bibinfo{person}{Jian Sun}.}
  \bibinfo{year}{2016}\natexlab{}.
\newblock \showarticletitle{Deep residual learning for image recognition}. In
  \bibinfo{booktitle}{\emph{Proceedings of the IEEE conference on computer
  vision and pattern recognition}}. \bibinfo{pages}{770--778}.
\newblock


\bibitem[\protect\citeauthoryear{He, Zhang, and Sun}{He et~al\mbox{.}}{2017}]%
        {he2017channel}
\bibfield{author}{\bibinfo{person}{Yihui He}, \bibinfo{person}{Xiangyu Zhang},
  {and} \bibinfo{person}{Jian Sun}.} \bibinfo{year}{2017}\natexlab{}.
\newblock \showarticletitle{Channel pruning for accelerating very deep neural
  networks}. In \bibinfo{booktitle}{\emph{Proceedings of the IEEE International
  Conference on Computer Vision}}. \bibinfo{pages}{1389--1397}.
\newblock


\bibitem[\protect\citeauthoryear{Hinton, Vinyals, and Dean}{Hinton
  et~al\mbox{.}}{2015}]%
        {hinton2015distilling}
\bibfield{author}{\bibinfo{person}{Geoffrey Hinton}, \bibinfo{person}{Oriol
  Vinyals}, {and} \bibinfo{person}{Jeff Dean}.}
  \bibinfo{year}{2015}\natexlab{}.
\newblock \showarticletitle{Distilling the knowledge in a neural network}.
\newblock \bibinfo{journal}{\emph{arXiv preprint arXiv:1503.02531}}
  (\bibinfo{year}{2015}).
\newblock


\bibitem[\protect\citeauthoryear{Hong, Sukumaran-Rajam, Bandyopadhyay, Kim,
  Kurt, Nisa, Sabhlok, {\c{C}}ataly{\"u}rek, Parthasarathy, and
  Sadayappan}{Hong et~al\mbox{.}}{2018}]%
        {hong2018efficient}
\bibfield{author}{\bibinfo{person}{Changwan Hong}, \bibinfo{person}{Aravind
  Sukumaran-Rajam}, \bibinfo{person}{Bortik Bandyopadhyay},
  \bibinfo{person}{Jinsung Kim}, \bibinfo{person}{S{\"u}reyya~Emre Kurt},
  \bibinfo{person}{Israt Nisa}, \bibinfo{person}{Shivani Sabhlok},
  \bibinfo{person}{{\"U}mit~V {\c{C}}ataly{\"u}rek},
  \bibinfo{person}{Srinivasan Parthasarathy}, {and} \bibinfo{person}{P
  Sadayappan}.} \bibinfo{year}{2018}\natexlab{}.
\newblock \showarticletitle{Efficient sparse-matrix multi-vector product on
  GPUs}. In \bibinfo{booktitle}{\emph{Proceedings of the 27th International
  Symposium on High-Performance Parallel and Distributed Computing}}. ACM,
  \bibinfo{pages}{66--79}.
\newblock


\bibitem[\protect\citeauthoryear{Hong, Sukumaran-Rajam, Nisa, Singh, and
  Sadayappan}{Hong et~al\mbox{.}}{2019}]%
        {hong2019adaptive}
\bibfield{author}{\bibinfo{person}{Changwan Hong}, \bibinfo{person}{Aravind
  Sukumaran-Rajam}, \bibinfo{person}{Israt Nisa}, \bibinfo{person}{Kunal
  Singh}, {and} \bibinfo{person}{P Sadayappan}.}
  \bibinfo{year}{2019}\natexlab{}.
\newblock \showarticletitle{Adaptive sparse tiling for sparse matrix
  multiplication}. In \bibinfo{booktitle}{\emph{Proceedings of the 24th
  Symposium on Principles and Practice of Parallel Programming}}. ACM,
  \bibinfo{pages}{300--314}.
\newblock


\bibitem[\protect\citeauthoryear{Howard, Zhu, Chen, Kalenichenko, Wang, Weyand,
  Andreetto, and Adam}{Howard et~al\mbox{.}}{2017}]%
        {howard2017mobilenets}
\bibfield{author}{\bibinfo{person}{Andrew~G Howard}, \bibinfo{person}{Menglong
  Zhu}, \bibinfo{person}{Bo Chen}, \bibinfo{person}{Dmitry Kalenichenko},
  \bibinfo{person}{Weijun Wang}, \bibinfo{person}{Tobias Weyand},
  \bibinfo{person}{Marco Andreetto}, {and} \bibinfo{person}{Hartwig Adam}.}
  \bibinfo{year}{2017}\natexlab{}.
\newblock \showarticletitle{Mobilenets: Efficient convolutional neural networks
  for mobile vision applications}.
\newblock \bibinfo{journal}{\emph{arXiv preprint arXiv:1704.04861}}
  (\bibinfo{year}{2017}).
\newblock


\bibitem[\protect\citeauthoryear{Jia, Maggioni, Smith, and Scarpazza}{Jia
  et~al\mbox{.}}{2019}]%
        {jia2019dissecting}
\bibfield{author}{\bibinfo{person}{Zhe Jia}, \bibinfo{person}{Marco Maggioni},
  \bibinfo{person}{Jeffrey Smith}, {and} \bibinfo{person}{Daniele~Paolo
  Scarpazza}.} \bibinfo{year}{2019}\natexlab{}.
\newblock \showarticletitle{Dissecting the NVidia Turing T4 GPU via
  microbenchmarking}.
\newblock \bibinfo{journal}{\emph{arXiv preprint arXiv:1903.07486}}
  (\bibinfo{year}{2019}).
\newblock


\bibitem[\protect\citeauthoryear{Jiang, Hong, and Agrawal}{Jiang
  et~al\mbox{.}}{2020}]%
        {jiang2020novel}
\bibfield{author}{\bibinfo{person}{Peng Jiang}, \bibinfo{person}{Changwan
  Hong}, {and} \bibinfo{person}{Gagan Agrawal}.}
  \bibinfo{year}{2020}\natexlab{}.
\newblock \showarticletitle{A novel data transformation and execution strategy
  for accelerating sparse matrix multiplication on GPUs}. In
  \bibinfo{booktitle}{\emph{Proceedings of the 25th ACM SIGPLAN Symposium on
  Principles and Practice of Parallel Programming}}. \bibinfo{pages}{376--388}.
\newblock


\bibitem[\protect\citeauthoryear{Lavin and Gray}{Lavin and Gray}{2016}]%
        {lavin2016fast}
\bibfield{author}{\bibinfo{person}{Andrew Lavin} {and} \bibinfo{person}{Scott
  Gray}.} \bibinfo{year}{2016}\natexlab{}.
\newblock \showarticletitle{Fast algorithms for convolutional neural networks}.
  In \bibinfo{booktitle}{\emph{Proceedings of the IEEE Conference on Computer
  Vision and Pattern Recognition}}. \bibinfo{pages}{4013--4021}.
\newblock


\bibitem[\protect\citeauthoryear{Lin, Ji, Li, Deng, and Li}{Lin
  et~al\mbox{.}}{2019}]%
        {lin2019toward}
\bibfield{author}{\bibinfo{person}{Shaohui Lin}, \bibinfo{person}{Rongrong Ji},
  \bibinfo{person}{Yuchao Li}, \bibinfo{person}{Cheng Deng}, {and}
  \bibinfo{person}{Xuelong Li}.} \bibinfo{year}{2019}\natexlab{}.
\newblock \showarticletitle{Toward Compact ConvNets via Structure-Sparsity
  Regularized Filter Pruning}.
\newblock \bibinfo{journal}{\emph{IEEE transactions on neural networks and
  learning systems}} (\bibinfo{year}{2019}).
\newblock


\bibitem[\protect\citeauthoryear{Louizos, Welling, and Kingma}{Louizos
  et~al\mbox{.}}{2017}]%
        {louizos2017learning}
\bibfield{author}{\bibinfo{person}{Christos Louizos}, \bibinfo{person}{Max
  Welling}, {and} \bibinfo{person}{Diederik~P Kingma}.}
  \bibinfo{year}{2017}\natexlab{}.
\newblock \showarticletitle{Learning Sparse Neural Networks through $ L\_0 $
  Regularization}.
\newblock \bibinfo{journal}{\emph{arXiv preprint arXiv:1712.01312}}
  (\bibinfo{year}{2017}).
\newblock


\bibitem[\protect\citeauthoryear{Molchanov, Ashukha, and Vetrov}{Molchanov
  et~al\mbox{.}}{2017}]%
        {molchanov2017variational}
\bibfield{author}{\bibinfo{person}{Dmitry Molchanov}, \bibinfo{person}{Arsenii
  Ashukha}, {and} \bibinfo{person}{Dmitry Vetrov}.}
  \bibinfo{year}{2017}\natexlab{}.
\newblock \showarticletitle{Variational dropout sparsifies deep neural
  networks}. In \bibinfo{booktitle}{\emph{Proceedings of the 34th International
  Conference on Machine Learning-Volume 70}}. JMLR. org,
  \bibinfo{pages}{2498--2507}.
\newblock


\bibitem[\protect\citeauthoryear{Narang}{Narang}{2016}]%
        {deepbench}
\bibfield{author}{\bibinfo{person}{Sharan Narang}.}
  \bibinfo{year}{2016}\natexlab{}.
\newblock \showarticletitle{Deepbench}.
\newblock  (\bibinfo{year}{2016}).
\newblock


\bibitem[\protect\citeauthoryear{Park, Li, Wen, Tang, Li, Chen, and Dubey}{Park
  et~al\mbox{.}}{2016}]%
        {park2016faster}
\bibfield{author}{\bibinfo{person}{Jongsoo Park}, \bibinfo{person}{Sheng Li},
  \bibinfo{person}{Wei Wen}, \bibinfo{person}{Ping Tak~Peter Tang},
  \bibinfo{person}{Hai Li}, \bibinfo{person}{Yiran Chen}, {and}
  \bibinfo{person}{Pradeep Dubey}.} \bibinfo{year}{2016}\natexlab{}.
\newblock \showarticletitle{Faster cnns with direct sparse convolutions and
  guided pruning}.
\newblock \bibinfo{journal}{\emph{arXiv preprint arXiv:1608.01409}}
  (\bibinfo{year}{2016}).
\newblock


\bibitem[\protect\citeauthoryear{Strout, Carter, Ferrante, and Kreaseck}{Strout
  et~al\mbox{.}}{2004}]%
        {strout2004sparse}
\bibfield{author}{\bibinfo{person}{Michelle~Mills Strout},
  \bibinfo{person}{Larry Carter}, \bibinfo{person}{Jeanne Ferrante}, {and}
  \bibinfo{person}{Barbara Kreaseck}.} \bibinfo{year}{2004}\natexlab{}.
\newblock \showarticletitle{Sparse tiling for stationary iterative methods}.
\newblock \bibinfo{journal}{\emph{The International Journal of High Performance
  Computing Applications}} \bibinfo{volume}{18}, \bibinfo{number}{1}
  (\bibinfo{year}{2004}), \bibinfo{pages}{95--113}.
\newblock


\bibitem[\protect\citeauthoryear{Strout, Hall, and Olschanowsky}{Strout
  et~al\mbox{.}}{2018}]%
        {strout2018sparse}
\bibfield{author}{\bibinfo{person}{Michelle~Mills Strout},
  \bibinfo{person}{Mary Hall}, {and} \bibinfo{person}{Catherine Olschanowsky}.}
  \bibinfo{year}{2018}\natexlab{}.
\newblock \showarticletitle{The sparse polyhedral framework: Composing
  compiler-generated inspector-executor code}.
\newblock \bibinfo{journal}{\emph{Proc. IEEE}} \bibinfo{volume}{106},
  \bibinfo{number}{11} (\bibinfo{year}{2018}), \bibinfo{pages}{1921--1934}.
\newblock


\bibitem[\protect\citeauthoryear{Tan and Le}{Tan and Le}{2019}]%
        {tan2019efficientnet}
\bibfield{author}{\bibinfo{person}{Mingxing Tan} {and} \bibinfo{person}{Quoc~V
  Le}.} \bibinfo{year}{2019}\natexlab{}.
\newblock \showarticletitle{EfficientNet: Rethinking Model Scaling for
  Convolutional Neural Networks}.
\newblock \bibinfo{journal}{\emph{arXiv preprint arXiv:1905.11946}}
  (\bibinfo{year}{2019}).
\newblock


\bibitem[\protect\citeauthoryear{Tillet and Cox}{Tillet and Cox}{2017}]%
        {tillet2017input}
\bibfield{author}{\bibinfo{person}{Philippe Tillet} {and}
  \bibinfo{person}{David Cox}.} \bibinfo{year}{2017}\natexlab{}.
\newblock \showarticletitle{Input-aware auto-tuning of compute-bound HPC
  kernels}. In \bibinfo{booktitle}{\emph{Proceedings of the International
  Conference for High Performance Computing, Networking, Storage and
  Analysis}}. ACM, \bibinfo{pages}{43}.
\newblock


\bibitem[\protect\citeauthoryear{Tillet, Kung, and Cox}{Tillet
  et~al\mbox{.}}{2019}]%
        {tillet2019triton}
\bibfield{author}{\bibinfo{person}{Philippe Tillet}, \bibinfo{person}{HT Kung},
  {and} \bibinfo{person}{David Cox}.} \bibinfo{year}{2019}\natexlab{}.
\newblock \showarticletitle{Triton: an intermediate language and compiler for
  tiled neural network computations}. In \bibinfo{booktitle}{\emph{Proceedings
  of the 3rd ACM SIGPLAN International Workshop on Machine Learning and
  Programming Languages}}. ACM, \bibinfo{pages}{10--19}.
\newblock


\bibitem[\protect\citeauthoryear{Vasilache, Zinenko, Theodoridis, Goyal,
  DeVito, Moses, Verdoolaege, Adams, and Cohen}{Vasilache
  et~al\mbox{.}}{2018}]%
        {vasilache2018tensor}
\bibfield{author}{\bibinfo{person}{Nicolas Vasilache},
  \bibinfo{person}{Oleksandr Zinenko}, \bibinfo{person}{Theodoros Theodoridis},
  \bibinfo{person}{Priya Goyal}, \bibinfo{person}{Zachary DeVito},
  \bibinfo{person}{William~S Moses}, \bibinfo{person}{Sven Verdoolaege},
  \bibinfo{person}{Andrew Adams}, {and} \bibinfo{person}{Albert Cohen}.}
  \bibinfo{year}{2018}\natexlab{}.
\newblock \showarticletitle{Tensor comprehensions: Framework-agnostic
  high-performance machine learning abstractions}.
\newblock \bibinfo{journal}{\emph{arXiv preprint arXiv:1802.04730}}
  (\bibinfo{year}{2018}).
\newblock


\bibitem[\protect\citeauthoryear{Vaswani, Shazeer, Parmar, Uszkoreit, Jones,
  Gomez, Kaiser, and Polosukhin}{Vaswani et~al\mbox{.}}{2017}]%
        {vaswani2017attention}
\bibfield{author}{\bibinfo{person}{Ashish Vaswani}, \bibinfo{person}{Noam
  Shazeer}, \bibinfo{person}{Niki Parmar}, \bibinfo{person}{Jakob Uszkoreit},
  \bibinfo{person}{Llion Jones}, \bibinfo{person}{Aidan~N Gomez},
  \bibinfo{person}{{\L}ukasz Kaiser}, {and} \bibinfo{person}{Illia
  Polosukhin}.} \bibinfo{year}{2017}\natexlab{}.
\newblock \showarticletitle{Attention is all you need}. In
  \bibinfo{booktitle}{\emph{Advances in neural information processing
  systems}}. \bibinfo{pages}{5998--6008}.
\newblock


\bibitem[\protect\citeauthoryear{Venkat, Hall, and Strout}{Venkat
  et~al\mbox{.}}{2015}]%
        {venkat2015loop}
\bibfield{author}{\bibinfo{person}{Anand Venkat}, \bibinfo{person}{Mary Hall},
  {and} \bibinfo{person}{Michelle Strout}.} \bibinfo{year}{2015}\natexlab{}.
\newblock \showarticletitle{Loop and data transformations for sparse matrix
  code}. In \bibinfo{booktitle}{\emph{ACM SIGPLAN Notices}},
  Vol.~\bibinfo{volume}{50}. ACM, \bibinfo{pages}{521--532}.
\newblock


\bibitem[\protect\citeauthoryear{Wang, Wohlwend, and Lei}{Wang
  et~al\mbox{.}}{2019}]%
        {wang2019structured}
\bibfield{author}{\bibinfo{person}{Ziheng Wang}, \bibinfo{person}{Jeremy
  Wohlwend}, {and} \bibinfo{person}{Tao Lei}.} \bibinfo{year}{2019}\natexlab{}.
\newblock \showarticletitle{Structured Pruning of Large Language Models}.
\newblock \bibinfo{journal}{\emph{arXiv preprint arXiv:1910.04732}}
  (\bibinfo{year}{2019}).
\newblock


\bibitem[\protect\citeauthoryear{Wen, He, Rajbhandari, Zhang, Wang, Liu, Hu,
  Chen, and Li}{Wen et~al\mbox{.}}{2017}]%
        {wen2017learning}
\bibfield{author}{\bibinfo{person}{Wei Wen}, \bibinfo{person}{Yuxiong He},
  \bibinfo{person}{Samyam Rajbhandari}, \bibinfo{person}{Minjia Zhang},
  \bibinfo{person}{Wenhan Wang}, \bibinfo{person}{Fang Liu},
  \bibinfo{person}{Bin Hu}, \bibinfo{person}{Yiran Chen}, {and}
  \bibinfo{person}{Hai Li}.} \bibinfo{year}{2017}\natexlab{}.
\newblock \showarticletitle{Learning intrinsic sparse structures within long
  short-term memory}.
\newblock \bibinfo{journal}{\emph{arXiv preprint arXiv:1709.05027}}
  (\bibinfo{year}{2017}).
\newblock


\bibitem[\protect\citeauthoryear{Whaley, Petitet, and Dongarra}{Whaley
  et~al\mbox{.}}{2001}]%
        {whaley2001automated}
\bibfield{author}{\bibinfo{person}{R~Clint Whaley}, \bibinfo{person}{Antoine
  Petitet}, {and} \bibinfo{person}{Jack~J Dongarra}.}
  \bibinfo{year}{2001}\natexlab{}.
\newblock \showarticletitle{Automated empirical optimizations of software and
  the ATLAS project}.
\newblock \bibinfo{journal}{\emph{Parallel computing}} \bibinfo{volume}{27},
  \bibinfo{number}{1-2} (\bibinfo{year}{2001}), \bibinfo{pages}{3--35}.
\newblock


\bibitem[\protect\citeauthoryear{Yang, Bulu{\c{c}}, and Owens}{Yang
  et~al\mbox{.}}{2018}]%
        {yang2018design}
\bibfield{author}{\bibinfo{person}{Carl Yang}, \bibinfo{person}{Ayd{\i}n
  Bulu{\c{c}}}, {and} \bibinfo{person}{John~D Owens}.}
  \bibinfo{year}{2018}\natexlab{}.
\newblock \showarticletitle{Design principles for sparse matrix multiplication
  on the GPU}. In \bibinfo{booktitle}{\emph{European Conference on Parallel
  Processing}}. Springer, \bibinfo{pages}{672--687}.
\newblock


\bibitem[\protect\citeauthoryear{Yao, Cao, Xiao, Zhang, and Nie}{Yao
  et~al\mbox{.}}{2019}]%
        {yao2019balanced}
\bibfield{author}{\bibinfo{person}{Zhuliang Yao}, \bibinfo{person}{Shijie Cao},
  \bibinfo{person}{Wencong Xiao}, \bibinfo{person}{Chen Zhang}, {and}
  \bibinfo{person}{Lanshun Nie}.} \bibinfo{year}{2019}\natexlab{}.
\newblock \showarticletitle{Balanced sparsity for efficient dnn inference on
  gpu}. In \bibinfo{booktitle}{\emph{Proceedings of the AAAI Conference on
  Artificial Intelligence}}, Vol.~\bibinfo{volume}{33}.
  \bibinfo{pages}{5676--5683}.
\newblock


\bibitem[\protect\citeauthoryear{Zhu, Zhang, Gu, and Xie}{Zhu
  et~al\mbox{.}}{2019}]%
        {zhu2019sparse}
\bibfield{author}{\bibinfo{person}{Maohua Zhu}, \bibinfo{person}{Tao Zhang},
  \bibinfo{person}{Zhenyu Gu}, {and} \bibinfo{person}{Yuan Xie}.}
  \bibinfo{year}{2019}\natexlab{}.
\newblock \showarticletitle{Sparse tensor core: Algorithm and hardware
  co-design for vector-wise sparse neural networks on modern gpus}. In
  \bibinfo{booktitle}{\emph{Proceedings of the 52nd Annual IEEE/ACM
  International Symposium on Microarchitecture}}. \bibinfo{pages}{359--371}.
\newblock


\end{thebibliography}

\end{document}